\documentclass[twocolumn,10pt]{article}

\usepackage{arxiv}

\usepackage[utf8]{inputenc} 
\usepackage[T1]{fontenc}    
\usepackage{url}            
\usepackage{graphicx}
\usepackage{booktabs}       
\usepackage{amsfonts}       
\usepackage{microtype}      

\usepackage{xcolor}
\usepackage{colortbl}
\usepackage{multirow}
\usepackage{enumitem}
\usepackage{tabularx}
\newcolumntype{Y}{>{\centering\arraybackslash}X}
\usepackage{tcolorbox}
\tcbuselibrary{skins}
\tcbuselibrary{breakable}
\usepackage{amsmath}
\usepackage{amssymb}
\usepackage{array}
\usepackage{makecell}

\usepackage{natbib}
\setcitestyle{aysep={}}
\let\cite\citep

\usepackage[colorlinks=true, allcolors=black]{hyperref}

\title{When Perception Collides with Priors: Diagnosing Modal Polarization in Video Understanding Models}

\author{
    \textbf{Yao Tong}\\
    Hangzhou Institute for Advanced Study\\
    University of Chinese Academy of Sciences\\
    Hangzhou, China\\
    \texttt{tongyao261@mails.ucas.ac.cn} \\
    \And
    \textbf{Jiaxin Qi}\\
    Computer Network Information Center\\
    Chinese Academy of Sciences\\
    Beijing, China\\
    \texttt{jxqi@cnic.cn} \\
    \authorrowbreak
    \textbf{Yan Cui}\\
    Hangzhou Institute for Advanced Study\\
    University of Chinese Academy of Sciences\\
    Hangzhou, China\\
    \texttt{cuiyan.ch@gmail.com} \\
    \And
    \textbf{Jinli Ou}\\
    Hangzhou Institute for Advanced Study\\
    University of Chinese Academy of Sciences\\
    Hangzhou, China\\
    \texttt{oujinli@zuaa.zju.edu.cn} \\
    \authorrowbreak
    \textbf{Yuhua Zheng}\\
    Hangzhou Institute for Advanced Study\\
    University of Chinese Academy of Sciences\\
    Hangzhou, China\\
    \texttt{zhengyuhua@ucas.ac.cn} \\
    \And
    \textbf{Jianqiang Huang}\thanks{Corresponding author.}\\
    Computer Network Information Center\\
    Chinese Academy of Sciences\\
    Beijing, China\\
    \texttt{jqhuang@cnic.cn} \\
}

\renewcommand{\headeright}{Preprint}
\renewcommand{\undertitle}{Preprint}
\renewcommand{\shorttitle}{When Perception Collides with Priors}

\hypersetup{
pdftitle={When Perception Collides with Priors: Diagnosing Modal Polarization in Video Understanding Models},
pdfsubject={cs.CV, cs.CL, cs.AI},
pdfauthor={Yao Tong, Jiaxin Qi, Yan Cui, Jinli Ou, Yuhua Zheng, Jianqiang Huang},
pdfkeywords={Multimodal Large Language Models, Video Understanding, Benchmark, Hallucination, Modal Polarization},
}

\begin{document}

\maketitle
\begin{abstract}
Multimodal Large Language Models (MLLMs) have achieved strong video comprehension, largely benefiting from extensive training on realistic video data. However, when video perception severely conflicts with internal priors or common sense, we identify a critical failure pattern termed Modal Polarization, where models fail to integrate modalities and instead bifurcate into two distinct extremes: Cognitive Inertia, where models rigidly adhere to priors despite contradictory visual evidence, and Perceptual Inertia, where they blindly trust video perception while ignoring logical inconsistencies.
To investigate this, we introduce TheaterBench, a benchmark designed to expose Modal Polarization using counter-intuitive scenarios that violate physical laws or common sense---situations rarely encountered in standard training distributions. Extensive experiments reveal that current state-of-the-art (SOTA) models exhibit severe Modal Polarization, and many de-hallucination methods merely rebalance the trade-off between Perceptual and Cognitive Inertia rather than fundamentally resolving the conflict.  This work serves as a foundational step for evaluating and enhancing the perceptual-cognitive equilibrium of future video understanding models.
\end{abstract}

\begin{figure*}[!t]
  \centering
  \includegraphics[width=0.90\linewidth]{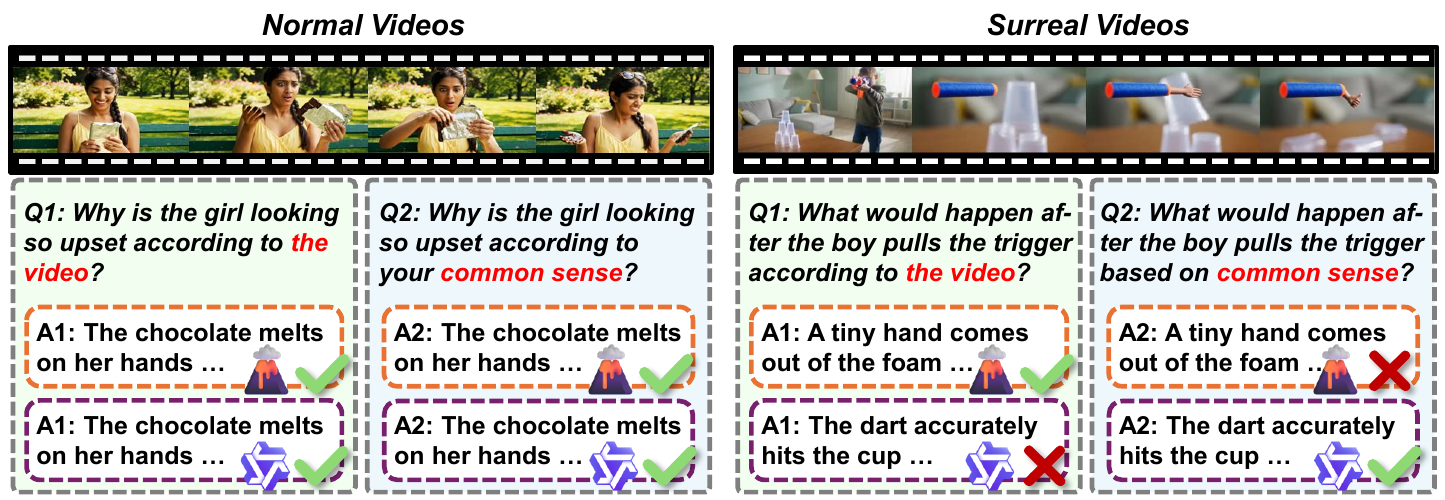}
  \caption{Overview of the TheaterBench evaluation framework using normal (left) and surreal (right) videos. The figure displays video frames alongside queries that specify either ``according to the video'' (green dashed box) or ``according to your common sense'' (blue dashed box). Different models exhibit distinct tendencies when facing surreal videos: models in the orange dashed box consistently answer based on the video content, while models in the plum dashed box always rely on common sense.}
  \label{fig:intro}
\end{figure*}

\begin{figure*}[t]
  \centering
  \includegraphics[width=0.95\linewidth]{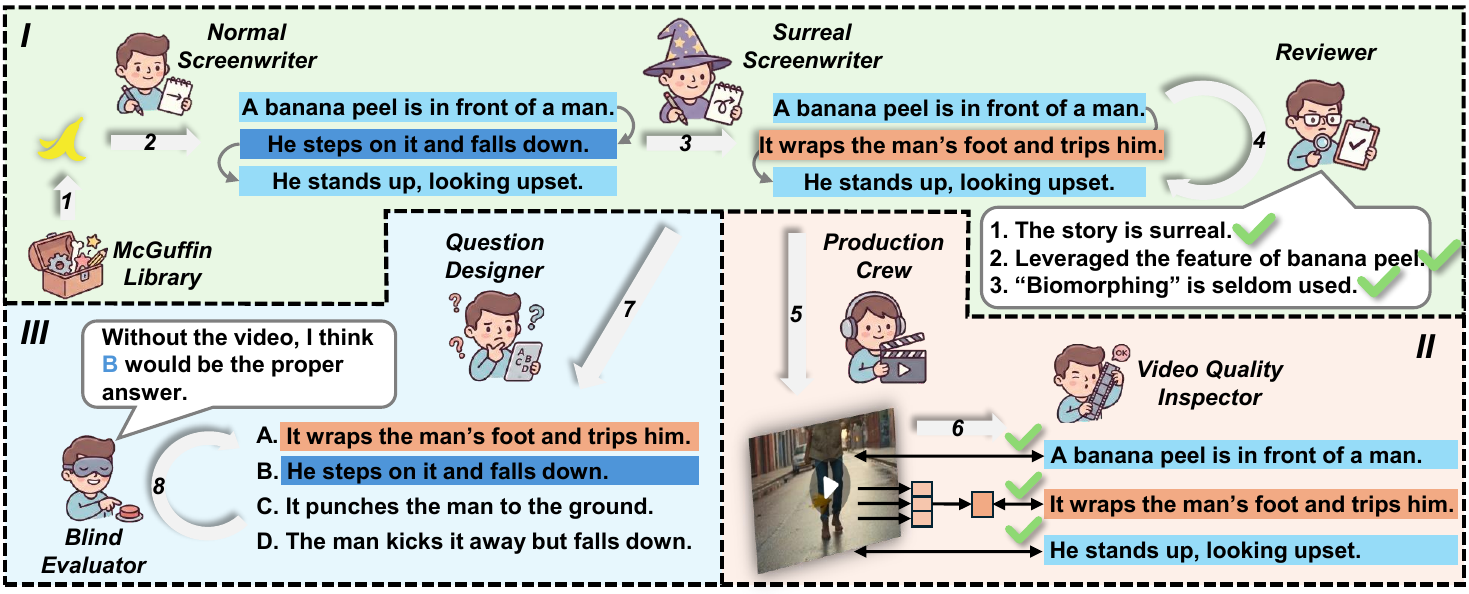}
  \caption{Overview of the TheaterBench dataset construction workflow. The pipeline consists of three phases: \textit{Phase I: Timeline Generation} (1. selecting an object, 2. scriptwriting, 3. rewriting, 4. quality vetting); \textit{Phase II: Video Generation} (5. video generation, 6. consistency check); and \textit{Phase III: QA Generation} (7. question formulation, 8. question filtering). The \textit{McGuffin} used here is \textit{banana peel}.}
  \label{fig:workflow}
\end{figure*}

\section{Introduction}
\label{sec:intro}

In recent years, video understanding has evolved beyond simple classification to complex tasks requiring the intricate fusion of video and language modalities \cite{chengVideoLLaMA2Advancing2024, liSurveyStateArt2025, linVideoLLaVALearningUnited2024, liuNVILAEfficientFrontier2025, maazVideoChatGPTDetailedVideo2024, wangInternVL35AdvancingOpenSource2025, zhangVideoLLaMA3Frontier2025}. Fundamentally, a robust video understanding model must synthesize visual evidence with textual constraints to derive precise answers \cite{coresLostTimeNew2025}. Benefiting from extensive pre-training on real-world video-text pairs \cite{chenPanda70MCaptioning70M2024, marcuLingoQAVisualQuestion2024, wangInternVidLargescaleVideoText2024}, existing SOTA models demonstrate impressive proficiency in these standard scenarios \cite{baiQwen3VLTechnicalReport2025, luoValleyVideoAssistant2025, xuQwen3OmniTechnicalReport2025, zhangLLaVAVideoVideoInstruction2025}. Since visual events in training distributions typically adhere to physical laws and match intuition, models can effortlessly synergize visual perception with linguistic guidance to execute downstream tasks.

Meanwhile, the rapid advancement of Video Generation Models (VGMs) \cite{timVideoGenerationModels, veo-teamVeo} has catalyzed a growing ecosystem of surreal and counterfactual video content. Creators increasingly leverage VGMs to produce physically impossible or fantastical scenarios---from magical transformations to impossible physics---to realize imaginative narratives unattainable in the real world. As such content proliferates across entertainment, education, and creative industries, it becomes critical for video understanding models to reliably interpret these counter-intuitive videos.

However, we discover that this fusion capability is severely compromised when visual content significantly contradicts the model's internal world knowledge.
In such counter-intuitive scenarios, for example, a tiny hand emerging from a flying foam dart (Figure~\ref{fig:intro}), models fail to maintain the equilibrium between vision and language.
We term this failure pattern \textit{Modal Polarization}.
Interestingly, models tend to bifurcate into two extremes based on which modality dominates: (a) \textit{Cognitive Inertia}: the model rigidly adheres to internal priors and hallucinates a normal outcome (e.g., ``a dart hitting the cup''), ignoring contradictory visual evidence; and (b) \textit{Perceptual Inertia}: the model blindly describes raw visual content without answering the inquiry or acknowledging the counterfactual nature of the instruction (e.g. ``according to your common sense'').
Crucially, this is a \textit{bidirectional} failure: while prior work has extensively studied the Cognitive Inertia direction under the umbrella of video hallucination, the Perceptual Inertia extreme---where models become completely enslaved by visual input---remains unexplored, as does the tradeoff between the two.
Furthermore, we observe that perception and cognition scores appear inversely correlated, highlighting the brittleness of current fusion mechanisms. This indicates that existing benchmarks, which primarily operate within the models' ``comfort zone'' of realistic videos, fail to distinguish whether a model genuinely fuses modalities or simply exploits the high correlation between visual content and intuitive priors \cite{liVideoHalluEvaluatingMitigating2025}.

To investigate this, we introduce \textbf{TheaterBench}, a benchmark constructed via a multi-stage pipeline to simulate counter-intuitive scenarios. Starting with diverse objects, we generate realistic scripts and rewrite their core developments into physically impossible events. These surreal events are ``Sandwiched'' between realistic beginning and ending segments, isolating the model's ability to reconcile cognitive expectations when visual evidence shifts to a surreal state. Scripts are then realized into videos using SOTA VGMs \cite{timVideoGenerationModels, veo-teamVeo} and undergo strict quality control to ensure visual fidelity.

Through extensive experiments, we reveal that even SOTA models suffer from severe Modal Polarization. Neither prompt engineering \cite{weiChainThoughtPrompting2022} nor inference-time interventions \cite{lengMitigatingObjectHallucinations2024, sunSmartSightMitigatingHallucination2025, zhangEventHallusionDiagnosingEvent2024} resolve this bottleneck; they merely shift the static bias from one extreme to the other. We further identify this polarization as a uniquely cross-modal, bidirectional failure that existing interventions cannot resolve.

Our contributions can be summarized as follows:
\begin{itemize}
    \item We introduce \textbf{TheaterBench}, a novel benchmark with counter-intuitive scenarios and a rigorous construction pipeline.
    \item We reveal that SOTA models suffer from severe Modal Polarization, and de-hallucination methods merely shift the static bias rather than resolving the conflict.
    \item We conduct a comprehensive analysis revealing that Modal Polarization is a uniquely cross-modal, bidirectional failure that existing interventions cannot resolve.
\end{itemize}

\section{Related Work}
\label{sec:related_work}

\noindent\textbf{Video Hallucination.}
Hallucination has evolved from a marginal artifact in LLMs \cite{maynezFaithfulnessFactualityAbstractive2020, jiSurveyHallucinationNatural2023, carliniExtractingTrainingData2021} to a pervasive challenge in MLLMs \cite{baiHallucinationMultimodalLarge2025, lengMitigatingObjectHallucinations2024, parkGLSimDetectingObject2025, zhangEventHallusionDiagnosingEvent2024}. While image hallucination research targets object existence \cite{renNoLanMitigatingObject2026, yangNulluMitigatingObject, zhouAnalyzingMitigatingObject2024}, video understanding introduces temporal vulnerabilities: sparse sampling creates information gaps that compel models to over-rely on language priors \cite{baeMASHVLMMitigatingActionScene2025, nieSlowFocusEnhancingFinegrained2026, wangVideoHallucerEvaluatingIntrinsic2024, zhangEventHallusionDiagnosingEvent2024}. This manifests in Temporal Attention Collapse (TAC) \cite{sunSmartSightMitigatingHallucination2025} and Action-Scene Hallucination \cite{baeMASHVLMMitigatingActionScene2025}, where models fabricate content based on static frames or background context. Mitigation methods such as VCD \cite{lengMitigatingObjectHallucinations2024}, TCD, and SmartSight \cite{sunSmartSightMitigatingHallucination2025} have been proposed, but these can weaken language-related capabilities \cite{ortuWhenSeeingOverrides2025}. Notably, existing video hallucination benchmarks predominantly evaluate unidirectional failures---whether models fabricate content absent from the video---within realistic scenarios, leaving the complementary problem of models suppressing valid visual evidence largely unexamined. Our work goes beyond prior failure modes: rather than studying unidirectional hallucination, we investigate the \textit{bidirectional} Modal Polarization where models collapse toward either perceptual or cognitive extremes.

\begin{figure*}[t]
  \centering
  \includegraphics[width=0.90\linewidth]{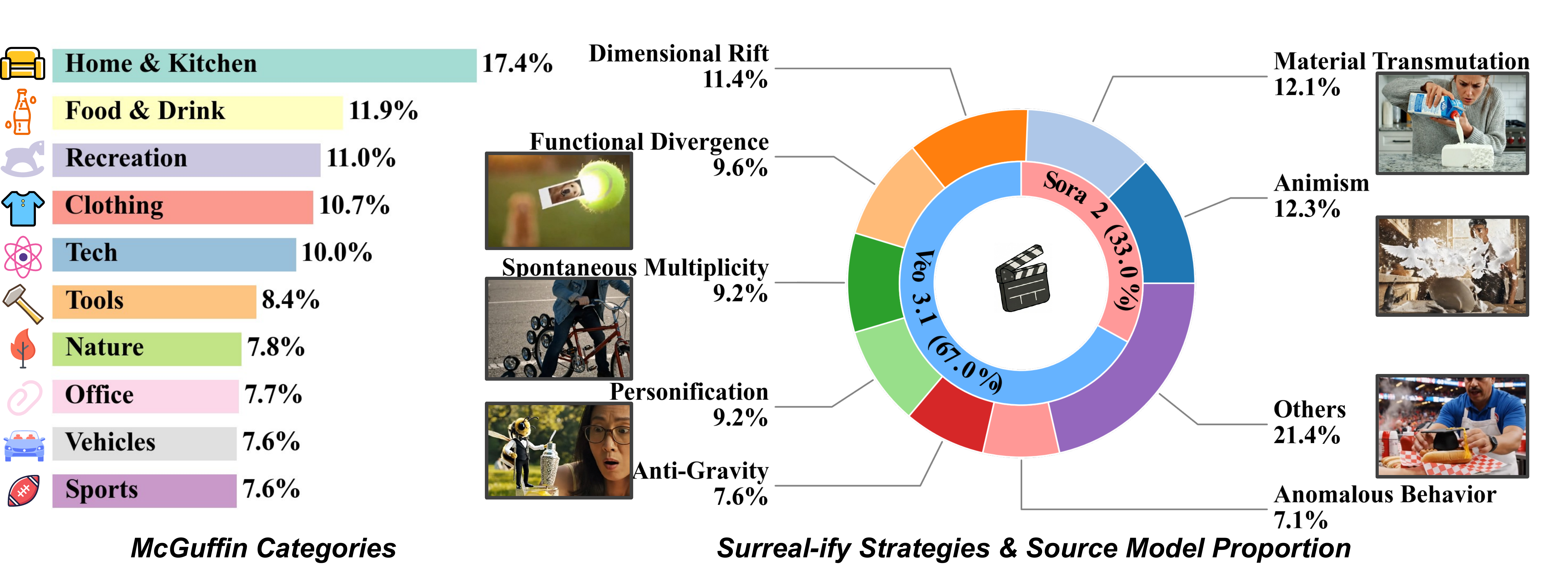}
  \caption{Data statistics of TheaterBench. Left: Frequency of McGuffin categories across the dataset. Right: Distribution of surreal-ify strategies and the proportion of source models (Sora 2 and Veo 3.1).}
  \label{fig:video_distri}
\end{figure*}

\noindent\textbf{Video Generation Models and Synthetic Video Datasets.}
Video Generation Models (VGMs) have transitioned from early GANs \cite{goodfellowGenerativeAdversarialNetworks2014, iglesiasSurveyGANsComputer2023} to Diffusion Transformers \cite{dhariwalDiffusionModelsBeat2021c, maEfficientDiffusionModels2025}, enabling Sora 2 \cite{timVideoGenerationModels} and Veo 3.1 \cite{veo-teamVeo} to produce high-fidelity videos. This has spurred new evaluation frameworks, such as VideoHallu's adversarial prompts for anomaly detection \cite{liVideoHalluEvaluatingMitigating2025} and the Video Reality Test for ASMR authenticity \cite{wangVideoRealityTest2025}. However, these benchmarks primarily assess whether models can detect generated artifacts or judge realism, rather than probing how models reason about content that deliberately violates physical laws. In contrast, we extend beyond anomaly detection to investigate whether MLLMs exhibit modality bias when presented with counter-intuitive stimuli, using a diverse McGuffin Library and automated classification to ensure dataset breadth.

\section{TheaterBench}
To systematically evaluate and quantify the Modal Polarization of MLLMs, we propose \textbf{TheaterBench}, a benchmark featuring counter-intuitive scenarios constructed through an autonomous agent-based pipeline.
Inspired by professional film production, we design a multi-stage workflow where LLMs \cite{deepseek-aiDeepSeekV3TechnicalReport2025}, Image-VLMs \cite{openaiGPT4TechnicalReport2024}, and VGMs \cite{timVideoGenerationModels, veo-teamVeo} collaborate as specialized agents---from screenwriting and directing to quality inspection.
The core of our methodology is the ``Sandwich'' temporal design, which provides a controlled environment to probe the conflict between visual perception and cognitive priors.

\subsection{Methodology}

Standalone surreal or normal videos are insufficient to simultaneously probe the limits of both perception and cognition. To effectively address this, we propose the ``\textit{Sandwich}'' temporal structure, where counter-intuitive content is injected between two realistic segments, thereby allowing surreal events to occur within a seemingly normal background:
\begin{itemize}
    \item Beginning Stage (from 0\% to 30\% of the total duration): Depicts a realistic setup involving a \textit{McGuffin} (e.g., a person holding a banana).
    \item Development Stage (from 30\% to 70\% of the total duration): The core surreal event occurs here, where the McGuffin's behavior or properties violate physical laws or common sense (e.g., the banana turns into a phone).
    \item Ending Stage (from 70\% to 100\% of the total duration): Returns to a realistic state consistent with the initial setup (e.g., the person puts away the object).
\end{itemize}

By creating this modal conflict within a single video, TheaterBench forces MLLMs to demonstrate genuine equilibrium in balancing visual evidence with world knowledge.

\subsection{Theater Production Team}
The construction of TheaterBench follows a rigorous pipeline analogous to a theater production.
The overall workflow is illustrated in Figure~\ref{fig:workflow}, involving three key phases: \textit{Phase I: Timeline Generation}, \textit{Phase II: Video Generation}, and \textit{Phase III: QA Generation}.

\noindent \textbf{Phase I: Timeline Generation}:
\begin{itemize}[nosep, leftmargin=1.5em]
    \item \raisebox{-0.25em}{\includegraphics[height=1.2em]{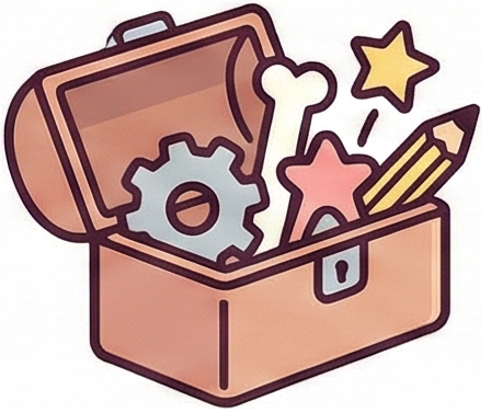}} \textit{McGuffin Library}: A curated collection of over 400 objects across various categories (e.g., tools, animals, food). A McGuffin serves as the central element that drives the plot.
    \item \raisebox{-0.25em}{\includegraphics[height=1.2em]{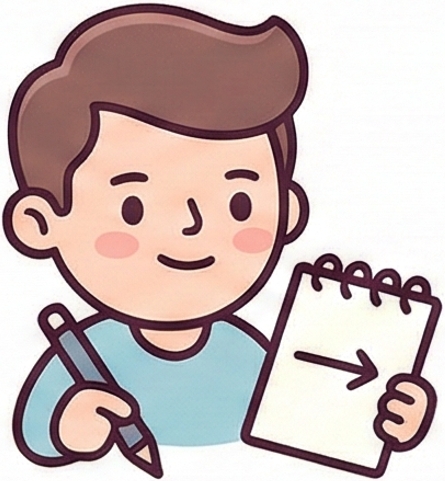}} \textit{Normal Screenwriter}: An LLM that writes a realistic timeline consisting of a beginning-development-ending sequence based on the McGuffin's natural properties.
    \item \raisebox{-0.25em}{\includegraphics[height=1.2em]{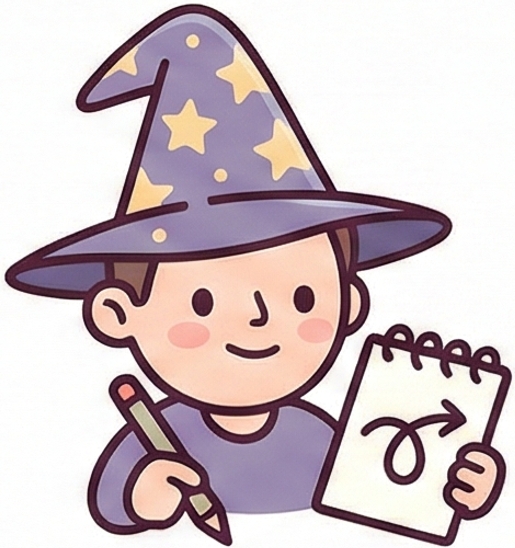}} \textit{Surreal Screenwriter}: An LLM that rewrites the development stage into counter-intuitive or physically impossible events (\textit{surreal-ify}) while maintaining temporal coherence with the realistic beginning and ending.
    \item \raisebox{-0.25em}{\includegraphics[height=1.2em]{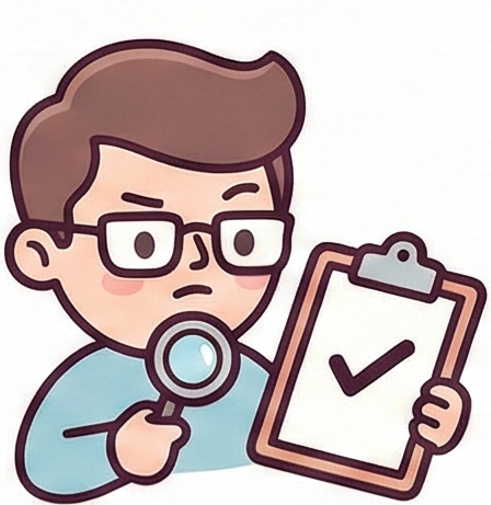}} \textit{Reviewer}: An LLM that vets scripts based on: (1) surreal intensity, (2) McGuffin-relevance, and (3) strategy diversity (prohibiting repetitive use of surreal-ify strategies like material transformation).
\end{itemize}

\noindent \textbf{Phase II: Video Generation}:
\begin{itemize}[nosep, leftmargin=1.5em]
    \item \raisebox{-0.25em}{\includegraphics[height=1.2em]{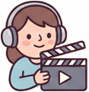}} \textit{Production Crew}: Utilizes SOTA VGMs (e.g., Sora 2, Veo 3.1) to synthesize videos from the reviewed surreal scripts, strictly adhering to the 30\%-40\%-30\% temporal distribution for the Sandwich structure.
    \item \raisebox{-0.25em}{\includegraphics[height=1.2em]{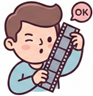}} \textit{Video Quality Inspector}: An Image-VLM that performs frame-by-frame semantic verification. It extracts frames from each stage and cross-references them with the original captions to ensure the visual content faithfully manifests the intended surreal conflict.
\end{itemize}

\noindent \textbf{Phase III: QA Generation}:
\begin{itemize}[nosep, leftmargin=1.5em]
    \item \raisebox{-0.25em}{\includegraphics[height=1.2em]{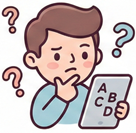}} \textit{Question Designer}: An LLM that formulates multiple-choice questions (MCQs).
    \item \raisebox{-0.25em}{\includegraphics[height=1.2em]{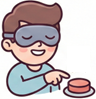}} \textit{Blind Evaluator}: An LLM that validates that the normal outcome remains the most intuitive choice when visual evidence is absent, ensuring the benchmark's cognitive rigor.
\end{itemize}

The automated pipeline acts as a strict filter, retaining only 8.6\% of initial candidates. To further ensure the benchmark's cognitive rigor, all 1,206 surviving video-QA pairs undergo a final human verification step: only samples that humans can answer correctly are retained, ultimately yielding the 984 entries in TheaterBench.

\begin{figure*}[t]
  \centering
  \includegraphics[width=0.95\linewidth]{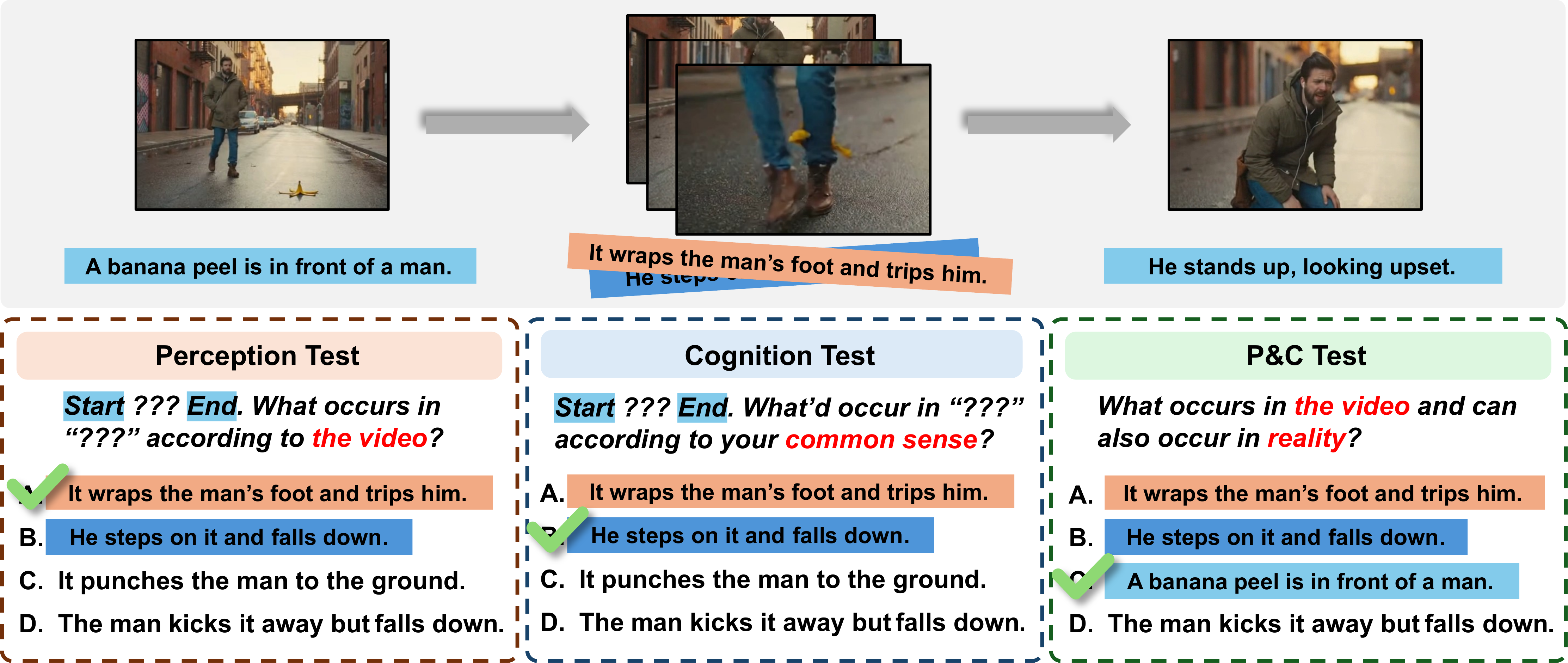}
  \caption{Examples of question variants in TheaterBench. We designed three types of tests: Perception (testing visual fidelity), Cognition (testing prior knowledge), and P\&C (Perception \& Cognition, testing the synergy between the two).}
  \label{fig:qa}
\end{figure*}

\subsection{Dataset Statistics}
TheaterBench contains 984 videos, including 325 Sora-generated 10-second videos and 659 Veo-generated 8-second videos, covering 10 categories of McGuffins, such as animals, objects, people, and places. Surreal timelines are generated with a diverse set of strategies, including animism, material transformation, and functional divergence, among others. Detailed data statistics are shown in Figure~\ref{fig:video_distri}.

\subsection{Question Variants}
To quantitatively evaluate a model's Modal Polarization, we designed three question types: (1) Perception Test, which asks the model to identify the surreal event based on visual evidence; (2) Cognition Test, which instructs the model to ignore the video and answer based on common sense; and (3) P\&C Test (Perception \& Cognition), which requires identifying a segment that occurs in both the video and reality. All tasks are formulated as four-option multiple-choice questions, as illustrated in Figure~\ref{fig:qa}.

\section{Experiment and Results}
\label{sec:experiment_and_results}

\subsection{Evaluation Metrics}

After evaluating all test items, we obtain the accuracies for the Perception Test, Cognition Test, and P\&C Test. We first define the corresponding scores $S_{P}$, $S_{C}$, and $S_{PC}$ as the accuracy multiplied by 100 to represent performance on a 100-point scale. Based on these, we introduce the following comprehensive metrics:

\begin{itemize}
    \item \textbf{Modal Competence Score (MCS)}: This score represents the overall capability of the model across different tasks. It is calculated as the arithmetic mean of the three task scores: $\text{MCS} = (S_{P} + S_{C} + S_{PC}) / 3$.

    \item \textbf{Modal Flexibility Score (MFS)}: This score measures the consistency of the model's performance on the same video across different task requirements. It is defined as the ratio of the number of videos where the model correctly handles all three task types ($N_{all}$) to the number of videos where it correctly handles at least one type ($N_{any}$), expressed as a percentage: $\text{MFS} = (N_{all} / N_{any}) \times 100$, where $N_{all} = \sum_{i} (c_{P}^{(i)} \cdot c_{C}^{(i)} \cdot c_{PC}^{(i)})$ and $N_{any} = \sum_{i} \mathbb{I}(c_{P}^{(i)} + c_{C}^{(i)} + c_{PC}^{(i)} > 0)$. The binary indicator $c_{Task}^{(i)} \in \{0, 1\}$ denotes whether the $i$-th video's task was correctly answered.

    \item \textbf{Modal Polarization Index (MPI)}: This index quantifies the model's performance imbalance between perception ($S_{P}$) and cognition ($S_{C}$) on surreal videos. It is calculated as the normalized difference between the two scores: $\text{MPI} = (S_{P} - S_{C}) / (S_{P} + S_{C}) \times 100$. The MPI reflects the model's modal preference: a positive value indicates a bias toward perception (\textit{Perceptual Inertia}), while a negative value indicates a bias toward cognition (\textit{Cognitive Inertia}). The larger the absolute value of MPI, the more pronounced this deviation becomes.
\end{itemize}

\subsection{Main Results}

\begin{table*}[t]
  \centering
  \begin{tabular}{l@{\hspace{10pt}}c@{\hspace{6pt}}c@{\hspace{10pt}}c@{\hspace{6pt}}c@{\hspace{8pt}}c@{\hspace{5pt}}c@{\hspace{5pt}}c@{\hspace{5pt}}c}
    \toprule
    \multirow{2}{*}{\textbf{Model}} & \multicolumn{2}{c}{$\mathbf{S_{P}}$} & \multicolumn{2}{c}{$\mathbf{S_{C}}$} & \multirow{2}{*}{\textbf{$\mathbf{S_{PC}}$}} & \multirow{2}{*}{\textbf{MFS}} & \multirow{2}{*}{\textbf{MCS}} & \multirow{2}{*}{\textbf{MPI}} \\
    \cmidrule(lr){2-3} \cmidrule(lr){4-5}
    & \textbf{Nor.} & \textbf{Sur.} & \textbf{Nor.} & \textbf{Sur.} & & & & \\
    \midrule
    \rowcolor{gray!15} \multicolumn{9}{c}{\textbf{Closed Source}} \\
    Gemini 2.5 Flash \cite{comaniciGemini25Pushing2025} & \textcolor{gray}{100.0} & 74.1 & \textcolor{gray}{100.0} & \textbf{96.8} & \underline{69.3} & \underline{54.7} & \underline{80.1} & $-$13.3 \\
    Gemini 3 Flash \cite{NewEraIntelligence2025} & \textcolor{gray}{93.5} & \textbf{87.7} & \textcolor{gray}{100.0} & \underline{81.8} & \textbf{75.5} & \textbf{55.0} & \textbf{81.7} & 3.5 \\
    Qwen3-VL-Plus \cite{baiQwen3VLTechnicalReport2025} & \textcolor{gray}{95.6} & 51.8 & \textcolor{gray}{98.9} & 70.9 & 24.7 & 6.6 & 49.1 & $-$15.6 \\
    Qwen-VL-Max \cite{baiQwen3VLTechnicalReport2025} & \textcolor{gray}{100.0} & \underline{76.6} & \textcolor{gray}{98.9} & 26.3 & 36.8 & 2.8 & 46.6 & 48.9 \\
    \rowcolor{gray!15} \multicolumn{9}{c}{\textbf{Open Source}} \\
    VideoLLaMA 3 \cite{zhangVideoLLaMA3Frontier2025} & \textcolor{gray}{98.9} & 55.6 & \textcolor{gray}{100.0} & 39.9 & 18.5 & 1.2 & 38.0 & 16.4 \\
    NVILA-Video \cite{liuNVILAEfficientFrontier2025} & \textcolor{gray}{98.3} & 50.8 & \textcolor{gray}{98.9} & 75.6 & 25.1 & 4.8 & 50.5 & $-$19.6 \\
    VideoChat-Flash \cite{liVideoChatFlashHierarchicalCompression2025} & \textcolor{gray}{95.6} & 70.3 & \textcolor{gray}{98.9} & 28.5 & 19.6 & 0.0 & 39.5 & 42.3 \\
    InternVL3.5 \cite{wangInternVL35AdvancingOpenSource2025} & \textcolor{gray}{96.7} & 28.8 & \textcolor{gray}{94.4} & 61.3 & 32.7 & 2.6 & 40.9 & $-$36.1 \\
    LLaVA-Video \cite{zhangLLaVAVideoVideoInstruction2025} & \textcolor{gray}{97.2} & 72.6 & \textcolor{gray}{100.0} & 27.7 & 10.8 & 1.0 & 37.0 & 44.8 \\
    Qwen3-Omni \cite{xuQwen3OmniTechnicalReport2025} & \textcolor{gray}{96.7} & 41.2 & \textcolor{gray}{94.4} & 72.4 & 16.5 & 4.0 & 43.4 & $-$27.5 \\
    Qwen3-VL-Flash \cite{baiQwen3VLTechnicalReport2025} & \textcolor{gray}{96.7} & 30.6 & \textcolor{gray}{96.7} & 70.9 & 51.1 & 5.4 & 50.9 & $-$39.7 \\
    \rowcolor{gray!15} \multicolumn{9}{c}{\textbf{Human}} \\
    Human & \textcolor{gray}{100.0} & 100.0 & \textcolor{gray}{100.0} & 100.0 & 100.0 & 100.0 & 100.0 & 0.0 \\
    \bottomrule
  \end{tabular}
  \caption{Quantitative results on TheaterBench. The best performance is highlighted in bold and the second best is \underline{underlined}. Nor. and Sur. denote Normal and Surreal videos, respectively. The data in \textcolor{gray}{gray} indicates results on normal videos.}
  \label{tab:main_results}
\end{table*}

The quantitative results in Table~\ref{tab:main_results} reveal several critical insights regarding the Modal Polarization of current MLLMs:

\textbf{Perceptual-Cognitive Synergy.}
The $S_{PC}$ measures the model's ability to simultaneously reconcile visual perception with cognitive reasoning in a single query. Most models struggle with this complex task, with scores often trailing significantly behind their performance in standalone perception or cognition tests. For example, LLaVA-Video achieves a perception score ($S_P$) of 72.6 but its $S_{PC}$ drops to only 10.8; similarly, Qwen3-VL-Plus scores 70.9 in cognition ($S_C$) but only 24.7 in $S_{PC}$. This highlights the difficulty of dynamic arbitration between modalities.

\textbf{Capacity Analysis.}
Gemini 3 Flash emerges as the top-performing model with a Modal Competence Score (MCS) of $81.7$, followed closely by Gemini 2.5 Flash at $80.1$. However, both lag significantly behind human performance ($100.0$). In the open-source community, Qwen3-VL-Flash ($50.9$) and Qwen3-VL-Plus ($49.1$) lead the pack, while LLaVA-Video ($37.0$) and VideoLLaMA 3 ($38.0$) show the lowest overall capability.

\textbf{Modal Polarization Analysis.}
We define a model as exhibiting Modal Polarization if its Modal Flexibility Score (MFS) is \textit{below 50} (e.g., VideoChat-Flash is 0.0 and LLaVA-Video is 1.0). This indicates that for the videos a model can answer at least partially, it fails to be fully consistent across all three tasks on more than half of them. Under this criterion, almost all evaluated models, except for the Gemini series, suffer from significant modal polarization. Within these polarized models, the Modal Polarization Index (MPI) further distinguishes the direction of their modal preference:
\begin{itemize}
    \item Perceptual Inertia (MPI $> 0$): Qwen-VL-Max ($48.9$), LLaVA-Video ($44.8$), and VideoLLaMA 3 ($16.4$). These models exhibit a ``blind vision'' behavior---they describe surreal details but lose their cognitive grounding, failing to distinguish between reality and surrealism.
    \item Cognitive Inertia (MPI $< 0$): Qwen3-VL-Flash ($-39.7$), InternVL3.5 ($-36.1$), NVILA-Video ($-19.6$), and Qwen3-VL-Plus ($-15.6$). These models suffer from ``knowledge hallucinations,'' where their internal common sense overpowers visual evidence, leading them to ignore surreal events entirely.
\end{itemize}

In summary, closed-source models indisputably lead across all metrics, including both overall capability (MCS) and modal flexibility (MFS), although a discernible gap remains when compared to human performance. Moreover, while some open-source models achieve competitive results in individual metrics such as $S_P$ or $S_C$, they significantly underperform relative to the Gemini series in MCS, and their MFS scores represent only a tiny fraction of those achieved by Gemini. This underscores that evaluating a model solely based on its video perception or common-sense knowledge is inadequate; rather, its ability to flexibly integrate these modalities under diverse task demands must be prioritized for a more comprehensive assessment.

\section{Analysis}
\label{sec:analysis}
In this section, we conduct a series of analyses to understand the root causes of the observed Modal Polarization issues. To represent the two ends of Modal Polarization, we select InternVL3.5 (Cognitive Inertia, negative MPI) and LLaVA-Video (Perceptual Inertia, positive MPI) as representative models for our investigations.

\subsection{Modal Polarization or Just a Mindset?}
To determine if the observed failures stem from a fixed cognitive ``mindset''---where models habitually ignore or obsess over anomalies regardless of the input type---we replace video inputs with their corresponding expanded captions. By keeping the information purely in the textual domain and ensuring comparable token lengths, we investigate whether this behavior persists when visual cues are absent. The results are summarized in Table~\ref{tab:caption_results}.

As shown in the table, both $S_{P}$ and $S_{C}$ are close to 100 in this setting (e.g., InternVL3.5 achieves $S_{P}$ of 99.2 and $S_{C}$ of 98.8). This indicates that the observed \textit{Modal Polarization} is indeed a failure to balance conflicting cues across different modalities, rather than a fixed cognitive mindset that would affect purely textual reasoning. When the information is presented entirely within the textual domain, models can effectively resolve conflicts and adhere to instructions.

\begin{table}[!t]
  \centering
  {\small
  \setlength{\tabcolsep}{3pt}
  \begin{tabular}{l@{\hspace{6pt}}c@{\hspace{6pt}}c@{\hspace{6pt}}c@{\hspace{4pt}}c@{\hspace{4pt}}c@{\hspace{4pt}}c}
    \toprule
    \textbf{Model} & $\mathbf{S_{P}}$ & $\mathbf{S_{C}}$ & $\mathbf{S_{PC}}$ & \textbf{MFS} & \textbf{MCS} & \textbf{MPI} \\
    \midrule
    InternVL3.5 (Video) & 28.8 & 61.3 & 32.7 & 2.6 & 40.9 & $-$36.1 \\
    InternVL3.5 (Caption) & 99.2 & 98.8 & 98.5 & 96.8 & 98.8 & 0.2 \\
    LLaVA-Video (Video) & 72.6 & 27.7 & 10.8 & 1.0 & 37.0 & 44.8 \\
    LLaVA-Video (Caption) & 98.5 & 99.1 & 98.2 & 96.2 & 98.6 & $-$0.3 \\
    \bottomrule
  \end{tabular}}
  \caption{Performance comparison between video inputs and expanded captions.}
  \label{tab:caption_results}
\end{table}

\begin{table}[!t]
  \centering
  {\small
  \setlength{\tabcolsep}{3pt}
  \begin{tabular}{l@{\hspace{4pt}}c@{\hspace{4pt}}c@{\hspace{4pt}}c@{\hspace{3pt}}c@{\hspace{3pt}}c@{\hspace{3pt}}c}
    \toprule
    \textbf{Model} & $\mathbf{S_{P}}$ & $\mathbf{S_{C}}$ & $\mathbf{S_{PC}}$ & \textbf{MFS} & \textbf{MCS} & \textbf{MPI} \\
    \midrule
    \rowcolor{gray!15} \multicolumn{7}{c}{\textbf{Base Models}} \\
    InternVL3.5 & 28.8 & 61.3 & 32.7 & 2.6 & 40.9 & $-$36.1 \\
    LLaVA-Video & 72.6 & 27.7 & 10.8 & 1.0 & 37.0 & 44.8 \\
    \rowcolor{gray!15} \multicolumn{7}{c}{\textbf{Prompt Engineering}} \\
    InternVL3.5+CoT & 13.1 & 70.4 & 37.8 & 3.6 & 40.4 & $-$68.6 \\
    InternVL3.5+RI & 28.2 & 73.0 & 21.3 & 3.0 & 40.8 & $-$44.3 \\
    LLaVA-Video+CoT & 60.6 & 36.7 & 8.9 & 0.7 & 35.4 & 24.6 \\
    LLaVA-Video+RI & 62.7 & 42.0 & 15.3 & 2.3 & 40.0 & 19.8 \\
    \rowcolor{gray!15} \multicolumn{7}{c}{\textbf{Inference Interventions}} \\
    InternVL3.5+VCD & 63.2 & 33.8 & 37.4 & 1.7 & 44.8 & 30.3 \\
    InternVL3.5+TCD & 53.4 & 44.7 & 30.8 & 2.3 & 43.0 & 8.9 \\
    InternVL3.5+SmartSight & 55.2 & 43.5 & 32.5 & 2.4 & 43.7 & 11.9 \\
    LLaVA-Video+VCD & 84.7 & 10.8 & 9.7 & 0.2 & 35.1 & 77.4 \\
    LLaVA-Video+TCD & 82.1 & 14.8 & 10.3 & 1.3 & 35.7 & 69.5 \\
    LLaVA-Video+SmartSight & 80.5 & 16.8 & 9.8 & 1.2 & 35.7 & 65.5 \\
    \hline
    InternVL3.5$-$VCD & 2.3 & 76.8 & 42.7 & 1.8 & 40.6 & $-$94.2 \\
    InternVL3.5$-$TCD & 13.4 & 71.8 & 43.7 & 1.2 & 43.0 & $-$68.5 \\
    InternVL3.5$-$SmartSight & 15.5 & 70.2 & 45.5 & 1.3 & 43.7 & $-$63.8 \\
    LLaVA-Video$-$VCD & 3.8 & 71.7 & 25.3 & 0.3 & 33.6 & $-$89.9 \\
    LLaVA-Video$-$TCD & 31.4 & 57.6 & 24.3 & 0.4 & 37.8 & $-$29.4 \\
    LLaVA-Video$-$SmartSight & 33.5 & 56.2 & 25.5 & 0.2 & 38.4 & $-$25.3 \\
    \bottomrule
  \end{tabular}}
  \caption{Comprehensive comparison of various intervention methods. ``+method'' indicates applying the method for de-hallucination, while ``$-$method'' indicates reversing the method to enhance priors.}
  \label{tab:analysis_results}
\end{table}

\begin{figure*}[t]
  \centering
  \includegraphics[width=0.85\linewidth]{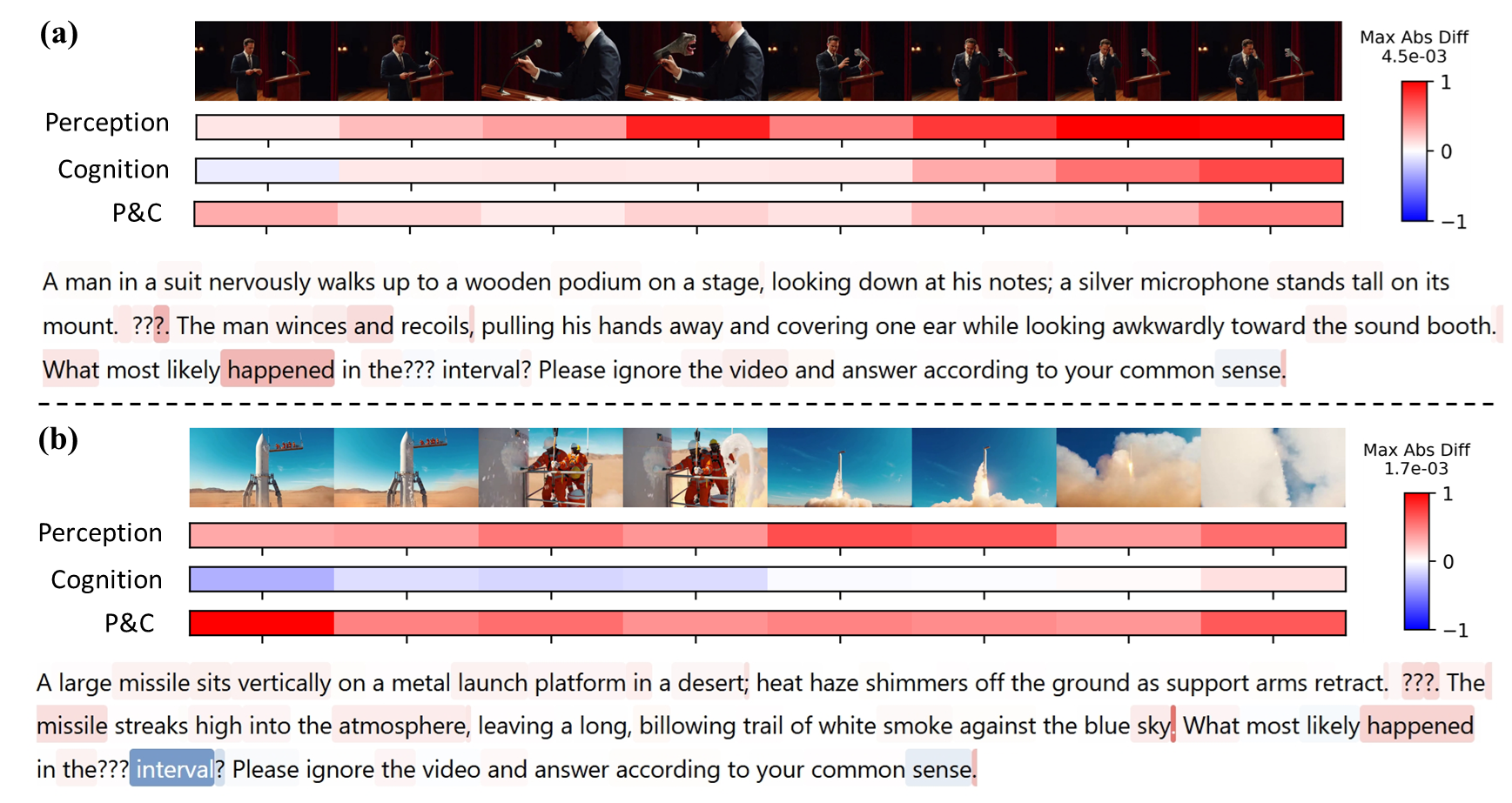}
  \caption{Visualization of attention score differences for middle layer of InternVL3.5 (SFT model minus base model). We present two examples: (a) a man recoils as a microphone transforms into a monster yelling at him, and (b) a person cleans a rocket during its launch. For each case, we visualize the frame-wise attention difference across Perception, Cognition, and P\&C tests, as well as the attention difference for question tokens specifically during the Cognition Test.}
  \label{fig:attn_diff}
\end{figure*}

\subsection{Is the Prompt to Blame?}

In our experiments, MLLMs often yield remarkably identical responses regardless of whether they are carefully instructed to attend to or ignore the video. Although original instructions such as "Please ignore the video and answer according to your common sense" are already sufficiently unambiguous, we consider the possibility that models might still overlook them: is this failure caused by the model's inability to follow subtle prompt variations? To investigate this, we systematically evaluate representative models using Chain-of-Thought (CoT) \cite{weiChainThoughtPrompting2022} prompting and Redundant Instructions (RI), where instructions are made exceptionally explicit or lengthy to ensure they are not overlooked. We aim to see if such enhanced reasoning or instruction following can lead to a synchronous improvement in $S_{P}$, $S_{C}$, and $S_{PC}$.

As shown in Table~\ref{tab:analysis_results}, both CoT and RI improve $S_{C}$ (e.g., +11.7 for InternVL3.5+RI and +14.3 for LLaVA-Video+RI) but fail to resolve the underlying conflict. This improvement does not stem from better instruction following; instead, the extended reasoning or redundant prompts likely distract the model from the video content, compelling a heavier reliance on common sense, as evidenced by the significant decline in $S_{P}$ (e.g., -15.7 for InternVL3.5+CoT and -12.0 for LLaVA-Video+CoT).

\subsection{What is the Cost of De-hallucination and Prior Enhancement?}
As discussed in Section~\ref{sec:related_work}, while numerous studies have focused on MLLM hallucinations and developed mitigation methods such as VCD \cite{lengMitigatingObjectHallucinations2024, sunSmartSightMitigatingHallucination2025, zhangEventHallusionDiagnosingEvent2024}, their potential side effects remain under-explored. Preliminary tests reveal that while these de-hallucination methods significantly bolster perceptual capabilities, they simultaneously lead to a substantial degradation in cognition. For instance, as shown in Table~\ref{tab:analysis_results}, applying VCD to InternVL3.5 increases $S_P$ from 28.8 to 63.2 but causes $S_C$ to drop from 61.3 to 33.8. Inspired by this, we reverse-engineered these methods to test the effects of hallucination enhancement (e.g., InternVL3.5$-$VCD). We found that while $S_C$ increases under these conditions, $S_P$ experiences a corresponding decline. For example, LLaVA-Video$-$VCD achieves an $S_C$ of 71.7, but its $S_P$ plummets to 3.8. This tradeoff between perception and cognition elucidates the core mechanism of Modal Polarization: models typically sacrifice performance in one modality to bolster the other, resulting in responses that consistently lean towards a single modality.

Moreover, the negligible change in MFS---which remains consistently below 3.0 across all settings---further confirms that these interventions are fundamentally a tradeoff. This demonstrates that if an intervention improves a model's accuracy on one type of query (e.g., the Perception Test), it invariably and predictably leads to incorrect responses on another (e.g., the Cognition Test), and vice-versa. Consequently, these methods fail to empower the model to correctly handle all three categories of questions simultaneously, highlighting a fundamental limitation in current inference-time interventions: they merely shift the static modality bias rather than fostering genuine dynamic flexibility.

\subsection{Further Analysis via SFT}
To further probe the nature of Modal Polarization, we apply Supervised Fine-Tuning (SFT) on counter-intuitive scenarios. Even modest SFT improves MFS for both InternVL3.5 (from 2.6 to 13.1) and LLaVA-Video (from 1.0 to 4.9), suggesting that exposure to surreal scenarios helps models develop modality flexibility.

To understand how SFT affects internal representations, we plot the attention-score delta (SFT $-$ base) for InternVL3.5 in Figure~\ref{fig:attn_diff}, focusing on middle layers where cross-modal associations form \cite{aflaloVLInterpreTInteractiveVisualization2022, huangPruneVidVisualToken2025, linBoostingMultimodalLarge2025}. During the Perception Test, the model attends more to frames with surreal cues. In the Cognition Test, attention increases toward prompt tokens describing realistic beginning/ending stages. In the P\&C Test, the model privileges segments aligning with both video and reality. These shifts suggest SFT enables dynamic reweighting between visual and textual evidence.

\section{Conclusion}
\label{sec:conclusion}
In this work, we identify the phenomenon of Modal Polarization in MLLMs, where models facing conflicts between visual perception and internal knowledge tend to collapse into either Cognitive Inertia or Perceptual Inertia. To systematically evaluate and quantify this deficiency, we introduce TheaterBench, a novel benchmark comprising counter-intuitive videos generated via a rigorous multi-agent pipeline. Our extensive experiments reveal that even current SOTA models struggle to reconcile these conflicting modalities. Furthermore, we identify this polarization as a uniquely cross-modal failure that persists even under strong prompt and intervention-based manipulations.

Despite these insights, our work has certain notable limitations. TheaterBench primarily focuses on counter-intuitive scenarios generated by VGMs, which may not fully encompass the diversity of real-world modality conflicts. Additionally, while we identify Modal Polarization as a cross-modal failure mode and attribute it to training data deficiencies, we do not propose a concrete mitigation strategy; developing effective and generalizable solutions remains an essential direction for future work.

\bibliographystyle{aaai2027}
\bibliography{TheaterBench}

\clearpage
\AppendicesPart
\appendix

\newtcolorbox{promptbox}[1]{
  colback=white,
  colframe=black,
  colbacktitle=gray!15!white,
  coltitle=black,
  fonttitle=\bfseries,
  title=#1,
  sharp corners,
  boxrule=0.5pt,
  left=4pt, right=4pt, top=4pt, bottom=4pt,
  breakable
}

\section{Dataset Generation Prompts}
We utilized a multi-agent pipeline to generate the video captions and scripts. The key prompts used by the Screenwriter agent are provided below.

\subsection{Normal Scene Generation}
The Normal Screenwriter agent is instructed to generate a realistic interaction timeline:

\begin{promptbox}{Prompt for Normal Scene Generation}
\scriptsize\ttfamily
"You must construct a realistic, normal interaction scene that is ENGAGING AND NOT BORING.
\begin{itemize}
    \item \textbf{Start (The Setup)}: Show the object and the subject. The subject can look at or approach the object, but MUST NOT TOUCH or interact with it yet.
    \item \textbf{Middle (The Interaction)}: This is where the normal interaction happens. The object should behave in a physically possible, expected way.
    \item \textbf{End (The Aftermath)}: The core action stops. Subject reacts to the normal interaction."
\end{itemize}
\end{promptbox}

\subsection{Surreal Scene Generation}
To create the Modal Polarization scenarios, the Surreal Screenwriter agent modifies the middle phase:

\begin{promptbox}{Prompt for Surreal Scene Generation}
\scriptsize\ttfamily
"You must modify ONLY the MIDDLE phase of an existing timeline to introduce a surreal/impossible/unusual element.
\begin{itemize}
    \item Keep start\_caption and end\_caption \textbf{IDENTICAL} to the original.
    \item Only modify middle\_caption to be surreal/impossible/unexpected/unusual.
    \item The twist should relate to the original normal interaction but defy expectations or reality."
\end{itemize}
\end{promptbox}

\subsubsection{Structure Auditor}
The Structure Auditor audits the generated timeline to enforce the [Setup $\to$ Twist $\to$ Reaction] structure.

\begin{promptbox}{Prompt for Structure Auditor}
\scriptsize\ttfamily
"You are a Timeline Structure Auditor. Enforce the [Setup -> Twist -> Reaction] format.

\textbf{\# STRUCTURE AUDIT:}
\begin{itemize}
    \item \textbf{Check Beginning}: Does the interaction (touching/action) start here? \textbf{IF YES -> REJECT.} (Must be setup only).
    \item \textbf{Check Middle}: Is there a TWIST? (Surreal/Unexpected/Unusual). If it's just normal action -> \textbf{REJECT}.
    \item \textbf{Check Ending}: Is there complex action? \textbf{IF YES -> REJECT.} (Must be simple reaction/staring).
    \item \textbf{Check Transition}: Is there continuity? If disjointed -> \textbf{REJECT}."
\end{itemize}
\end{promptbox}

\subsubsection{Scenario Classifier}
The Scenario Classifier categorizes the transformation from normal to surreal (Surreal-ification). Additionally, it monitors the cumulative distribution to ensure a balanced dataset; specifically, it rejects samples belonging to the top three most frequent scenarios. Furthermore, to enhance diversity, it rejects any scenario that has already been associated with the same McGuffin (the central object of the interaction).

\begin{promptbox}{Prompt for Scenario Classifier}
\scriptsize\ttfamily
"You are a Scenario Classifier. Your task is to analyze the change between a Normal Interaction and a Surreal/Unexpected Interaction.

\textbf{\# CLASSIFICATION PROTOCOL}
\begin{itemize}
    \item \textbf{Analyze the Change}: How did the object behavior change from Normal to Surreal?
    \item \textbf{Match Scenario}: Compare the change against the criteria of each defined scenario.
    \item \textbf{Select Best Fit}: Choose the scenario that best describes the surreal element."
\end{itemize}
\end{promptbox}

\section{Video Quality Assurance Prompts}
To ensure the quality and coherence of the generated videos, we employed a VLM-based Video Quality Inspector agent. This automated verification pipeline resulted in a video retention rate of 36.8\% (rejecting 63.2\% of generated videos), with 91.3\% of the rejections attributed to the middle phase (surrealism verification).

\subsection{Semantic Consistency Check}
To verify that the generated videos align with the intended surreal scenarios, we used a multi-step verification process involving frame description, story aggregation, and final verification.

\subsubsection{Step 1: Frame Description}
The following prompt was used to generate detailed descriptions for each extracted frame:

\begin{promptbox}{Prompt for Frame Description}
\scriptsize\ttfamily
"You are an expert image describer. Describe the main action and content of this image in detail."
\end{promptbox}

\subsubsection{Step 2: Story Aggregation}
The individual frame descriptions were then aggregated into a coherent story using the following prompt:

\begin{promptbox}{Prompt for Story Aggregation}
\scriptsize\ttfamily
"You are an expert storyteller. Summarize the following sequence of image descriptions into a concise, coherent story."
\end{promptbox}

\subsubsection{Step 3: Surrealism Verification}
The core prompt for the Surrealism Verification agent is:

\begin{promptbox}{Prompt for Video Quality Inspector}
\scriptsize\ttfamily
"Analyze the story for two things:
\begin{enumerate}
    \item Is the story counter-intuitive or surreal? (Does it defy physics or common sense in a surprising way?)
    \item Does it match the original middle caption?
\end{enumerate}

Scenarios:
\begin{itemize}
    \item 1. The story is NOT counter-intuitive/surreal. (Video is unqualified/boring).
    \item 2. The story IS counter-intuitive/surreal, BUT it differs significantly from the original middle caption.
    \item 3. The story IS counter-intuitive/surreal AND matches the original middle caption."
\end{itemize}
\end{promptbox}

\subsection{Normal Video Evaluation}
For the normal videos (control group), we used another verification prompt to ensure adherence to physical laws and the original caption:

\begin{promptbox}{Prompt for Normal Video Evaluation}
\scriptsize\ttfamily
"Analyze the story for:
\begin{enumerate}
    \item Does the video story accurately reflect the key events and actions described in the original caption?
\end{enumerate}
Since this is a 'Normal' mode evaluation, the video should simply follow the prompt without any surreal or unexpected twists unless explicitly requested by the caption."
\end{promptbox}

\begin{figure*}[t]
    \centering
    \includegraphics[width=0.8\linewidth]{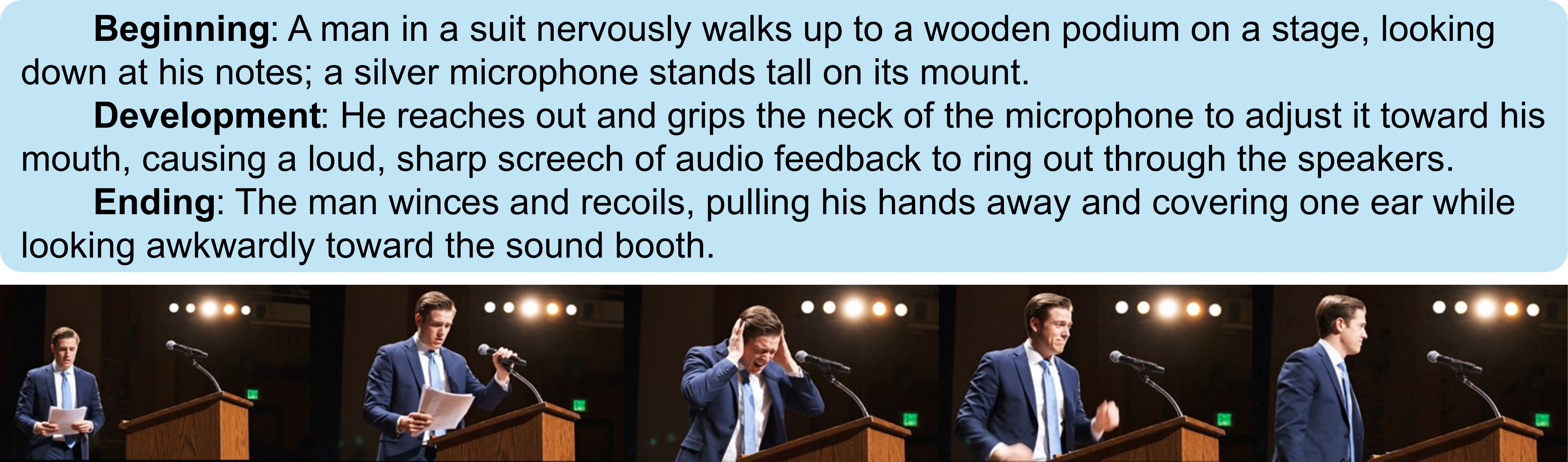}
    \caption{Example of a Normal Mode video successfully passed evaluation.}
    \label{fig:normal_eval}
\end{figure*}

\subsection{Beginning/Ending Frame Verification}
Given the simplicity of the beginning (setup) and ending (aftermath) stages, we employed a direct image-text matching verification to ensure the video begins and concludes as intended.

\begin{promptbox}{Prompt for Beginning/Ending Frame Verification}
\scriptsize\ttfamily
"You are a Video Consistency Evaluator.
Task: Verify if the provided video frame matches the scene description.

Input:
1. Scene Description: \{caption\}
2. Image: \{frame\}

Checklist:
- Are the main objects/subjects present?
- Is the described state/action clearly visible?

Score (0-10): Rate the visual alignment.
- 10: Perfect match.
- 0: Completely unrelated."
\end{promptbox}

\subsection{Scoring and Selection Mechanism}
The video selection process involves two steps:

\subsubsection{1. Qualification Filter}
A video is considered valid only if it meets the passing criteria for ALL three stages: \textit{beginning stage} (Visual alignment score $\ge 6.0$), \textit{middle stage} (Classified as Scenario 2 or Scenario 3; Scenario 1 is disqualified), and \textit{ending stage} (Visual alignment score $\ge 6.0$).

\subsubsection{2. Final Selection}
From the qualified candidates, we select the best video based on the highest Final Score, calculated as a weighted average (Weights: Start=0.2, Middle=0.6, End=0.2):
\small
\[
\text{Final Score} = 0.2 \times \text{Beg. Score} + 0.6 \times \text{Mid. Score} + 0.2 \times \text{End. Score}
\]
Where Middle Scores are assigned as follows: 10.0 for Scenario 3 (Surreal \& Matches) or Normal Mode Pass; 8.0 for Scenario 2 (Surreal but Differs); and 0.0 for Scenario 1 (Not Surreal) or Fail.

\begin{figure*}[t]
    \centering
    \includegraphics[width=0.9\linewidth]{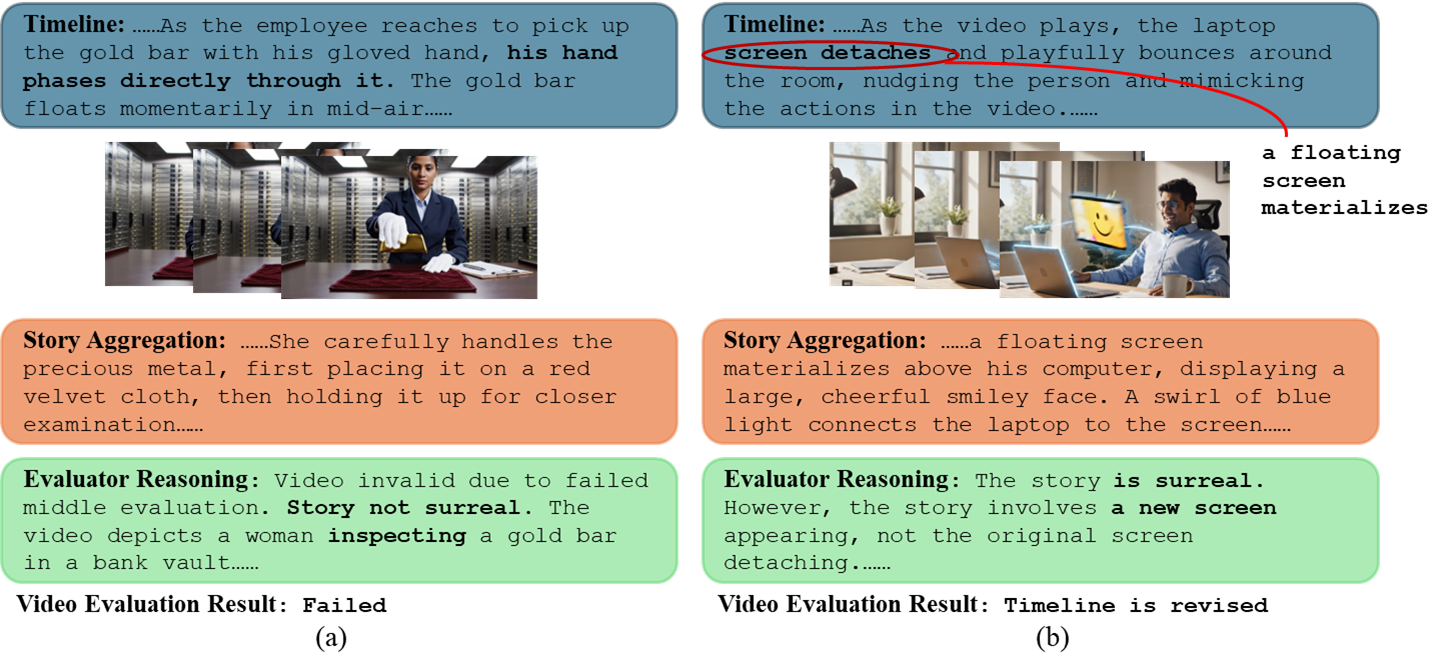}
    \caption{Example of Video Quality Control. Left: Scenario 1 (Rejection): The VLM-based evaluator correctly identifies that the generated video fails to depict the surreal element (phasing hand) described in the prompt, classifying it as a normal interaction (Scenario 1) and thus rejecting it. Right: Scenario 2 (Revision): The generated video contains a surreal element (floating screen) but it differs from the prompt (screen detaches). The system classifies it as Scenario 2 and triggers a revision of the timeline caption to match the generated video content.}
    \label{fig:video_eval}
\end{figure*}

\section{Question Generation Prompts}
We utilized dedicated agents to construct the multiple-choice questions (MCQs) for both the main Video QA benchmark and the auxiliary normal video benchmark (control group).

\subsection{Question Variant Details}
\label{appendix:question_details}

\subsubsection{Perception Test}
This type of question focuses exclusively on perception, and contradicts common sense. The question provides captions for the beginning and ending stages of a video and asks participants to describe what happened in between based on the video content. Among the options, the correct answer is the caption describing the developmental stage of the surreal timeline. A strong distractor is the caption for the developmental stage of the normal timeline, while the other two distractors are generated by the question designers.

\subsubsection{Cognition Test}
In contrast to the Perception Test, this type relies solely on common sense and necessitates effectively disregarding visual perception. Similar to the Perception Test, captions for the beginning and ending stages of a video are provided, but participants are instructed to ignore the video and answer based on common sense. The options are identical to those in the Perception Test, but the correct answer is the caption for the developmental stage of the normal timeline.

\subsubsection{P\&C Test}
This type of question simultaneously evaluates perceptual and cognitive capabilities. It directly asks participants: ``Which of the following segments occurs in the video and can also occur in reality?''. The options include:
\begin{itemize}
    \item Correct Option (Reality \& Video): A randomly selected caption from the start or end stage. These segments occur in both the video and reality.
    \item Distractor 1 (Reality \& !Video): The caption for the developmental stage of the normal timeline. This is possible in reality but does not occur in the current (surreal) video.
    \item Distractor 2 (!Reality \& Video): The caption for the middle part of the surreal timeline. This occurs in the video but is impossible in reality.
    \item Distractor 3 (!Reality \& !Video): A randomly selected distractor from the generated MCQ options, which neither occurs in the video nor in reality.
\end{itemize}

\subsection{Video Question Generation (Question Designer)}
For the main video understanding task, we constructed MCQs where the model must identify the correct event in the ``Middle'' interval. To prevent models from answering correctly based solely on text priors (Blind Guessing), we employed a ``Quizmaster'' agent (the Question Designer) to generate highly plausible distractors.

\begin{promptbox}{Prompt for Question Designer (Video QA)}
\scriptsize\ttfamily
"You are an Exam Creator's assistant.
Task: Generate 2 additional distractor options for a 4-option MCQ.
GOAL: Anti-Blind-Guessing.

\# EXISTING INFORMATION:
- Question prompt: \{question\}
- One existing option (distractor): \{normal\_middle\_caption\}
- One existing option (correct answer): \{correct\_middle\_caption\}

\# DISTRACTOR GENERATION STRATEGY (CRITICAL):
\begin{itemize}
    \item Generate 2 more highly misleading distractors that make it impossible for Video VLMs relying solely on text priors to answer correctly.
    \item Your generated options should NOT be identical to the existing distractor or correct answer.
    \item Options should be plausible but incorrect based on normal expectations.
\end{itemize}

\# STYLE \& TONE MATCHING (CRITICAL):
\begin{enumerate}
    \item \textbf{MATCH LENGTH}: Your options must have roughly the same word count as the existing options.
    \item \textbf{MATCH DETAIL LEVEL}: Maintain similar level of detail in your descriptions.
\end{enumerate}

OUTPUT JSON (RAW ONLY, NO MARKDOWN): \{ 'distractor1': '...', 'distractor2': '...' \}"
\end{promptbox}

\subsection{MQA Quality Control}
To rigorously filter out questions that can be answered without visual information, we employed a Blind Evaluator agent that attempts to guess the answer using only the text options. This automated filtering process retained 89.1\% of the generated questions. Specifically, a question is retained only if the Blind Evaluator, relying on text priors, selects the "normal" (common-sense) option rather than the correct "surreal" option. This ensures that the correct answer cannot be deduced from text alone and requires visual perception to override the common-sense prior.

\section{Details for Analysis}
\label{sec:appendix_analysis}

\subsection{Instructions for Prompt Analysis}
\label{appendix:prompts_analysis}

In the Analysis section of the main paper, we evaluated the impact of redundant instructions (RI) on model performance. The comparison between the default instructions and the redundant instructions is detailed below:

\begin{table*}[t]
  \centering
  {\small
  \begin{tabular}{m{2.5cm}@{\hspace{2pt}}m{4.5cm}@{\hspace{3pt}}m{9cm}}
    \toprule
    \textbf{Test Type} & \textbf{Default Instruction} & \textbf{Redundant Instruction} \\
    \midrule
    \textbf{Perception} & \ttfamily "Please answer according to the video." & \ttfamily "Please carefully analyze the provided video and identify the event that occurred during the '???' interval. Ensure your answer is strictly based on the visual evidence in the video." \\
    \addlinespace
    \textbf{Cognition} & \ttfamily "Please ignore the video and answer according to your common sense." & \ttfamily "Please ignore the video content and rely solely on your common sense and knowledge of physical laws. Select the option that describes the most realistic and logically consistent event for the '???' interval." \\
    \addlinespace
    \textbf{P\&C} & \ttfamily "Which of the following segments occurs in the video and can also occur in reality?" & \ttfamily "Please carefully analyze the provided video and compare the events with your knowledge of reality. Identify which of the following segments actually occurs in the video and is also physically possible in the real world." \\
    \bottomrule
  \end{tabular}}
  \caption{Comparison of default and redundant instructions.}
  \label{tab:instructions_comparison}
\end{table*}

\subsection{Timeline-to-Caption Expansion Prompts}
\label{sec:appendix_caption_expansion}
We employed DeepSeek-V3 to expand the generated timeline into detailed captions. Since video inputs naturally contain a vast amount of redundant tokens, we aimed to preserve this characteristic in the textual modality. We explicitly instructed the model to focus on describing environmental details and other redundant information, while keeping the caption for the middle phase relatively short. The system prompt used for this expansion is provided below:

\begin{promptbox}{Prompt for Timeline Expansion}
\scriptsize\ttfamily
"You are a text processing assistant. Your task is to rewrite the video timeline segments according to specific rules.

Input will be a JSON object with 'start', 'middle', and 'end'.

Rules:
1. \textbf{Start Caption:} EXPAND it significantly into a detailed description of the setting, atmosphere, and initial state (approx. 800-1000 words). Focus on environmental details, lighting, textures, and sensory information to create a rich context.
2. \textbf{Middle Caption:} REWRITE it to be more natural and engaging, but KEEP THE LENGTH SIMILAR to the original (approx. 30-50 words). Do not expand the action sequence significantly, just paraphrase it effectively.
3. \textbf{End Caption:} EXPAND it significantly into a detailed description of the reaction or aftermath (approx. 800-1000 words). Focus on the lingering atmosphere, emotions, and the static state of the world after the event.

\textbf{Goal:} The final combined text MUST be at least 2000 tokens long.

Output Format:
Return a strict JSON object with keys: "expanded\_start", "rewritten\_middle", "expanded\_end"."
\end{promptbox}

\begin{figure*}[t]
    \centering
    \includegraphics[width=\linewidth]{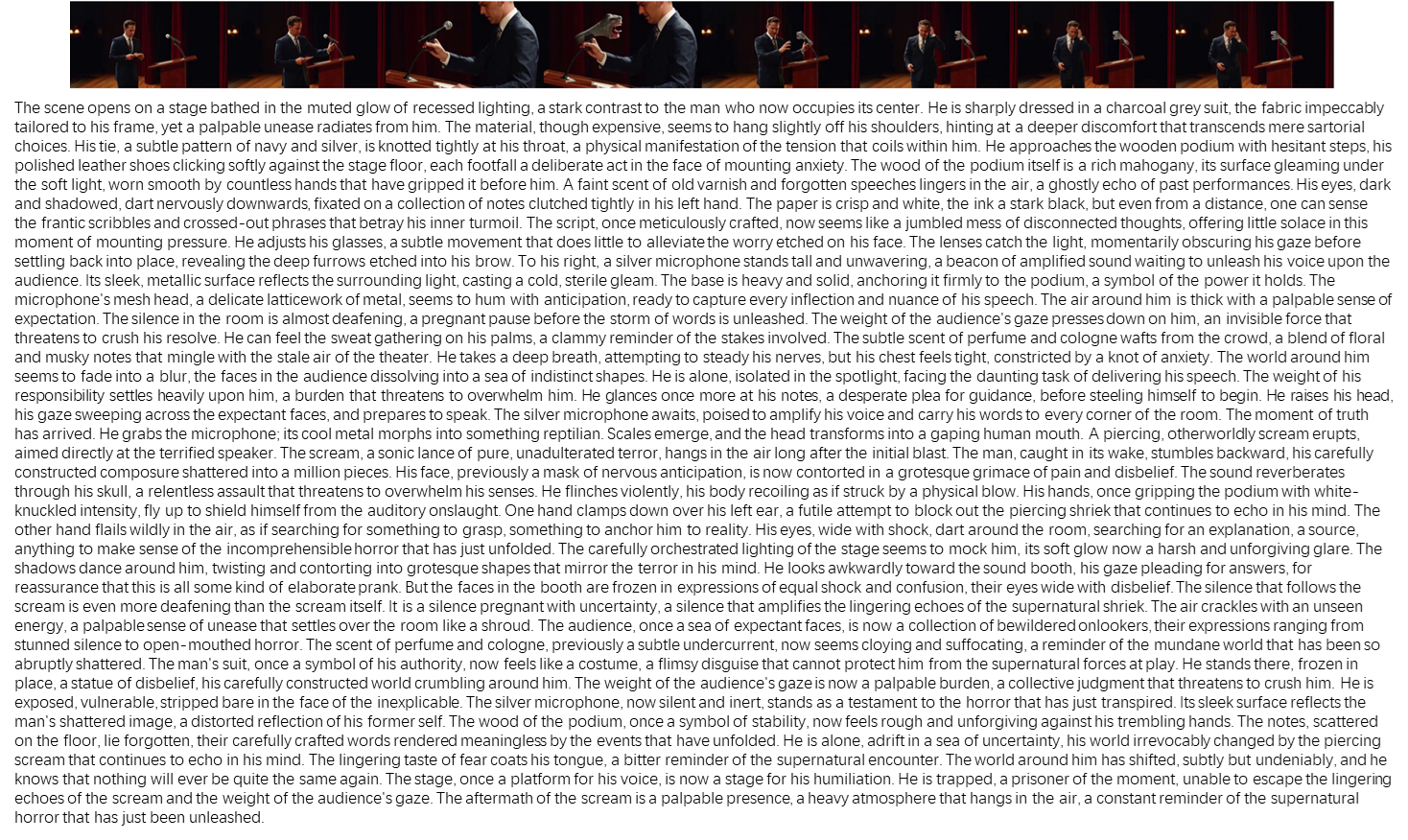}
    \caption{An example of a video sequence and its corresponding expanded caption. To eliminate the potential influence of token length differences between video and text inputs, we significantly expanded the original short captions into detailed descriptions, matching the information density of the video.}
    \label{fig:expanded_caption_example}
\end{figure*}

\subsection{Implementation Details of Interventions}
\label{appendix:intervention_details}

In the Analysis section of the main paper, we utilized several inference-time interventions to analyze the impact of visual and language priors.

\subsubsection{Visual Contrastive Decoding (VCD)}
Visual Contrastive Decoding (VCD) is an inference-time intervention method designed to mitigate hallucinations by contrasting the output distribution of the original input against a distorted baseline. In our experiments, we employ VCD and its inverse variant to analyze and manipulate the balance between visual perception and language priors. The adjusted logits are calculated using the following formula:

\begin{equation}
    \text{logits}_{\text{final}} = (1 + \alpha) \cdot \text{logits}_{\text{original}} - \alpha \cdot \text{logits}_{\text{distorted}}
\end{equation}

Here, $\text{logits}_{\text{original}}$ denotes the model's output logits derived from the normal video input, which contains the complete visual information. Conversely, $\text{logits}_{\text{distorted}}$ represents the output logits obtained from a black video input (constructed using \texttt{torch.zeros\_like}), thereby isolating the pure language priors or potential hallucinations.

The coefficient $\alpha$ controls the strength and direction of the adjustment. Specifically, for +VCD, we set $\alpha = 1$ to suppress language priors, whereas for $-$VCD, we set $\alpha = -1$ to enhance them.

\subsubsection{Temporal Contrastive Decoding (TCD)}
Temporal Contrastive Decoding (TCD) utilizes a ``weakened video'' containing only a few key frames as a contrastive baseline to enhance or suppress the model's focus on temporal details.

The adjustment formula is defined as:
\begin{equation}
    \text{logits}_{\text{final}} = (1 + \alpha) \cdot \text{logits}_{\text{original}} - \alpha \cdot \text{logits}_{\text{downsampled}}
\end{equation}

Here, $\text{logits}_{\text{original}}$ corresponds to the input with the full frame count (8 frames), preserving rich temporal information. In contrast, $\text{logits}_{\text{downsampled}}$ is derived from a downsampled video retaining only a few frames (2 frames), where temporal information is significantly lost while the main static content remains.

The coefficient $\alpha$ is set to $0.5$ for +TCD and $-0.5$ for $-$TCD. Furthermore, for +TCD, to prevent over-penalization, an adaptive thresholding mechanism is introduced:
\begin{equation}
    \text{threshold} = \beta \times \max(\text{logits}_{\text{original}})
\end{equation}
If the adjusted $\text{logits}_{\text{final}}$ falls below this threshold (with $\beta = 0.1$), the corresponding token is masked.

\subsubsection{SmartSight}
SmartSight is a method designed to mitigate or enhance hallucinations based on the Temporal Attention Collapse (TAC) score. In our experiments, we set the number of candidate responses $N=10$, the sampling temperature to $0.7$, the continuous steps for Visual Attention Vanishing (VAV) to $10$, the VAV attention ratio threshold to $1.2$, and the video input resolution to $392$.

The core mechanism of SmartSight relies on evaluating the Temporal Attention Collapse (TAC) score, denoted as $\mathcal{S}(y)$, which quantifies whether a model's attention disproportionately collapses onto isolated frames or semantically trivial segments. Importantly, a lower TAC score $\mathcal{S}(y)$ indicates a higher degree of hallucination.

For a given video and query, we independently sample $N$ candidate responses in parallel. To evaluate and control the hallucination severity without completing the full generation for all candidates, we compute the TAC score at the Visual Attention Vanishing (VAV) point---the step where the model's attention to visual tokens substantially diminishes.

Depending on our experimental objectives, we apply the following two distinct selection strategies:

\noindent +SmartSight (Mitigating Hallucination) Following the original implementation, we select the candidate response that exhibits the least attention collapse to suppress hallucinations. Specifically, we select the candidate $n^*$ that maximizes the TAC score:
\begin{equation}
    n^* = \arg\max_n (\mathcal{S}(y_{1:j_{vav}(y^n)}^n))
\end{equation}
By terminating sub-optimal candidates and continuing the generation of the response with the highest $\mathcal{S}(y)$, we effectively mitigate perception hallucinations while maintaining computational efficiency.

\noindent $-$SmartSight (Enhancing Hallucination) To intentionally induce or maximize hallucinations for our specific experimental analysis, we reverse the selection criterion. Since a lower $\mathcal{S}(y)$ corresponds to a sharper concentration of attention on isolated or trivial visual elements (higher hallucination), we select the candidate $n^*$ that minimizes the TAC score:
\begin{equation}
    n^* = \arg\min_n (\mathcal{S}(y_{1:j_{vav}(y^n)}^n))
\end{equation}

\subsection{Supervised Fine-Tuning Setup}
To investigate the impact of fine-tuning on the cognitive flexibility of MLLMs, we conducted Supervised Fine-Tuning (SFT) experiments.

\noindent Data Construction. We constructed a balanced dataset comprising Perception Test, Cognition Test, and P\&C Test samples in a 2:1:1 ratio. Unlike the multiple-choice format used in evaluation, we reformatted the data for SFT by using the correct option as the target label. The final dataset split consists of 400 training samples and 2,546 test samples.

\noindent Layer Configuration. We categorize the model layers into three groups: Low, Mid, and High. For InternVL3.5 (32 layers), these correspond to layers 0--9, 10--20, and 21--31, respectively. For LLaVA-Video (28 layers), these correspond to layers 0--8, 9--18, and 19--27, respectively. The ``All'' configuration denotes fine-tuning across all available layers.

\begin{table}[t]
  \centering
  {\small
  \setlength{\tabcolsep}{3pt}
  \begin{tabular}{l@{\hspace{6pt}}c@{\hspace{6pt}}c@{\hspace{6pt}}c@{\hspace{4pt}}c@{\hspace{4pt}}c@{\hspace{4pt}}c}
    \toprule
    \textbf{Model} & $\mathbf{S_{P}}$ & $\mathbf{S_{C}}$ & $\mathbf{S_{PC}}$ & \textbf{MFS} & \textbf{MCS} & \textbf{MPI} \\
    \midrule
    InternVL3.5 & 28.8 & 61.3 & 32.7 & 2.6 & 40.9 & $-$36.1 \\
    InternVL3.5+Low & 40.2 & 76.6 & 47.7 & 14.4 & 54.8 & $-$31.2 \\
    InternVL3.5+Mid & 42.7 & 69.9 & 40.8 & 13.1 & 51.1 & $-$24.1 \\
    InternVL3.5+High & 39.4 & 76.1 & 45.0 & 9.3 & 53.5 & $-$31.7 \\
    InternVL3.5+All & 35.6 & 73.3 & 44.6 & 9.2 & 51.2 & $-$34.5 \\
    LLaVA-Video & 72.6 & 27.7 & 10.8 & 1.0 & 37.0 & 44.8 \\
    LLaVA-Video+Low & 71.6 & 29.4 & 14.7 & 0.0 & 38.5 & 41.8 \\
    LLaVA-Video+Mid & 74.8 & 32.7 & 19.6 & 4.9 & 42.4 & 39.1 \\
    LLaVA-Video+High & 74.3 & 28.7 & 22.8 & 5.4 & 41.9 & 44.2 \\
    LLaVA-Video+All & 66.0 & 30.1 & 21.4 & 1.1 & 39.2 & 37.4 \\
    \bottomrule
  \end{tabular}}
  \caption{Performance of base models and their SFT versions across different layer configurations on TheaterBench.}
  \label{tab:sft_results}
\end{table}

\noindent Training Configuration. We employed LoRA (Low-Rank Adaptation) for efficient fine-tuning. The LoRA adapters were injected into all linear layers of the attention mechanism ($W_{q}, W_{k}, W_{v}, W_{o}$) and the feed-forward network ($W_{1}, W_{2}, W_{3}$). The rank $r$ was set to 16 with a scaling factor $\alpha$ of 32 and a dropout rate of 0.05. The model was trained for 1 epoch using the AdamW optimizer with a global batch size of 16 and a maximum sequence length of 4096 tokens. We used a cosine annealing scheduler with an initial learning rate of $4 \times 10^{-5}$ and a warmup ratio of 0.03. Weight decay was set to 0.01, and the training was conducted in BF16 precision. All fine-tuning experiments were performed on a single NVIDIA A800 PCIe GPU (80GB).

\noindent \textbf{Analysis of SFT Results.} Several key observations can be made from the results in Table~\ref{tab:sft_results}.
First, fine-tuning the middle layers (Mid) yields the most balanced performance for both models. For InternVL3.5, the Mid configuration achieves the highest $S_P$ (42.7) and the best MPI ($-$24.1), indicating it most effectively mitigates the model's inherent ``hallucination of normalcy.'' Similarly, for LLaVA-Video, the Mid configuration reaches peak scores in both $S_P$ (74.8) and $S_C$ (32.7). This suggests that middle layers---where cross-modal semantic associations are primarily formed---are the most critical regions for fostering modal flexibility.
Second, SFT significantly mitigates Modal Polarization. The MFS of InternVL3.5 increases from 2.6 to over 13.0, while LLaVA-Video sees an improvement from 1.0 to 4.9. This confirms that direct supervision on surreal scenarios effectively empowers models to dynamically reweight visual and textual evidence.
Third, full-layer fine-tuning (All) is not always superior to targeted local fine-tuning. For instance, InternVL3.5+All underperforms all local configurations in $S_P$, suggesting that unconstrained global parameter updates may lead to a sub-optimal convergence compared to focusing on specific functional layers.

\section{Detailed Experimental Results}
\label{sec:detailed_results}
We provide the detailed breakdown of error patterns and distractor analysis in this section.

\subsection{Error Pattern Analysis}
As shown in Table~\ref{tab:cognition_error}, when models with high positive MPI (e.g., Qwen-VL-Max, VideoChat-Flash) fail Cognition tasks, over $91\%$ of their errors involve choosing the surreal timeline, suggesting a ``knowledge neglect'' where the model is so fixated on visual perception that it fails to invoke common-sense reasoning to identify the event as impossible.
Conversely, Table~\ref{tab:perception_error} reveals that when models with significant negative MPI (e.g., Qwen3-VL-Flash, InternVL3.5) fail Perception tasks, over $93\%$ of errors consist of choosing the normal timeline despite visual evidence to the contrary. This indicates a ``hallucination of normalcy'' where internal priors override visual perception.

\subsection{Distractor Analysis}
To further investigate the nature of the errors, we analyze the distribution of the three types of distractors in the incorrect answers during the P\&C Test, as shown in Table~\ref{tab:pc_distractors}. The results clearly indicate that when models fail, they overwhelmingly gravitate towards Distractor 1---content that is consistent with real-world common sense but absent from the video.

\begin{table}[t]
\centering
{\small
\begin{tabular}{l@{\hspace{10pt}}c@{\hspace{10pt}}c}
\toprule
\textbf{Model} & $\mathbf{S_{P}}$ & \textbf{Ratio} \\
\midrule
Qwen3-Omni & 41.1 & 93.8\% \\
InternVL3.5 & 28.8 & 97.0\% \\
\bottomrule
\end{tabular}}
\caption{Analysis of perception errors choosing normal timeline (ratio).}
\label{tab:perception_error}
\end{table}

\begin{table}[t]
\centering
{\small
\begin{tabular}{l@{\hspace{15pt}}c@{\hspace{15pt}}c}
\toprule
\textbf{Model} & $\mathbf{S_{C}}$ & \textbf{Ratio} \\
\midrule
VideoChat-Flash & 28.5 & 91.2\% \\
LLaVA-Video & 27.7 & 94.5\% \\
\bottomrule
\end{tabular}}
\caption{Analysis of cognition errors choosing surreal timeline (ratio).}
\label{tab:cognition_error}
\end{table}

\begin{table}[t]
  \centering
  {\small
  \begin{tabular}{lccc}
    \toprule
    \textbf{Model} & \textbf{D1 (\%)} & \textbf{D2 (\%)} & \textbf{D3 (\%)} \\
    \midrule
    Gemini 3 Flash & 84.6 & 15.4 & 0.0 \\
    Gemini 2.5 Flash & 100.0 & 0.0 & 0.0 \\
    Qwen3-VL-Plus & 90.5 & 8.4 & 1.1 \\
    Qwen3-VL-Flash & 85.5 & 6.5 & 6.1 \\
    Qwen-VL-Max & 58.9 & 41.1 & 0.0 \\
    InternVL3.5 & 14.7 & 30.8 & 54.6 \\
    LLaVA-Video & 47.8 & 49.4 & 2.8 \\
    \bottomrule
  \end{tabular}}
  \caption{Distribution of distractors in incorrect answers during the P\&C Test. Distractor 1 (D1) represents content consistent with reality but absent from the video (Reality \& !Video). Distractor 2 (D2) represents content in the video that contradicts reality (!Reality \& Video). Distractor 3 (D3) represents content that is neither in reality nor the video (!Reality \& !Video).}
  \label{tab:pc_distractors}
\end{table}

\section{Further Discussion}
\label{sec:appendix_discussion}

\subsection{Conceptual Boundaries}
We clarify the relationship between Modal Polarization and two closely related concepts from prior work:

\noindent\textbf{Modal Polarization vs. Video Hallucination.}
Video hallucination describes models fabricating content absent from or inconsistent with the visual input, typically treated as a unidirectional failure where language priors overwrite perception. Modal Polarization subsumes this direction as Cognitive Inertia but crucially reveals its symmetric counterpart, Perceptual Inertia, where models become enslaved by visual content to the point of ignoring explicit textual instructions and common sense. More importantly, we show these two extremes exhibit a strict trade-off: improving one necessarily degrades the other on the same video instance. This bidirectional competition is not captured by existing hallucination frameworks, which examine each failure mode in isolation.

\noindent\textbf{Modal Polarization vs. Instruction Following Failure.}
Instruction following failures typically refer to models misunderstanding or ignoring task directives due to ambiguity, complexity, or insufficient training. Perceptual Inertia may superficially resemble such failures---e.g., a model ignoring ``answer according to common sense''---but the root cause is distinct. In our setting, the instruction is unambiguous, and models demonstrate full compliance when evaluated on normal videos. The failure arises specifically when strong visual stimuli override the model's ability to execute an otherwise trivial instruction-following behavior. This suggests a stimulus-driven override mechanism rather than a linguistic comprehension deficit.

\subsection{On the Use of VGM-Generated Surreal Videos}
TheaterBench relies entirely on videos synthesized by Video Generation Models (Sora 2 and Veo 3.1). We address two potential concerns regarding this design choice.

\noindent\textbf{Are model failures due to VGM artifacts rather than genuine modality conflict?}
VGMs are known to exhibit systematic artifacts such as spatial inconsistency, temporal discontinuity, and object flickering. A natural concern is whether the reported Modal Polarization stems from degraded visual signals rather than a true perceptual-cognitive conflict. We argue against this interpretation on two grounds. First, as detailed in the human verification study, human participants achieved 100\% accuracy on sampled questions and reported the tasks as straightforward. If VGM artifacts rendered the visual evidence genuinely ambiguous or unreliable, human annotators would similarly struggle to distinguish the surreal event from the normal outcome. The fact that humans perform flawlessly while even SOTA models exhibit severe polarization indicates that the failure originates from the model's inability to reconcile conflicting modalities, not from insufficient visual quality. Second, our Video Quality Inspector (an Image-VLM performing frame-by-frame verification) explicitly filters out videos with structural incoherence, ensuring that the surreal event is clearly manifested and temporally stable.

\noindent\textbf{VGM artifacts are themselves an inevitable challenge.}
Even if some degree of Modal Polarization could be attributed to VGM imperfections, this does not diminish the practical relevance of our findings. As AI-generated video content proliferates across social media, advertising, and entertainment, MLLMs will encounter not only counter-intuitive scenarios but also the artifacts inherent to current VGMs---spatial glitches, temporal jumps, and phantom objects. A robust video understanding system must navigate both the semantic conflict of surreal content and the signal noise of generative imperfections. TheaterBench captures this dual challenge as it exists in practice, rather than treating it as a confound to be eliminated.

\subsection{Broader Impact}
\noindent\textbf{Positive Impact.}
By exposing the Modal Polarization failure mode in current MLLMs, this work aims to promote the development of more robust and trustworthy video understanding systems. As AI-generated content becomes increasingly prevalent, the ability of MLLMs to reliably distinguish between physically plausible and impossible events---and to flexibly adapt their reasoning based on task demands---will be critical in applications ranging from content moderation to AI-assisted education. TheaterBench provides a diagnostic tool for measuring progress toward this goal.

\noindent\textbf{Potential Risks.}
The characterization of Modal Polarization could theoretically be exploited to craft adversarial inputs that deliberately trigger Cognitive Inertia or Perceptual Inertia in deployed models. For instance, malicious actors might design videos that force a model into blind trust of visual content, bypassing its common-sense reasoning. We believe that disclosing these failure modes is nonetheless net positive: awareness is a prerequisite for defense, and the benchmark provides a standardized way for the community to evaluate and mitigate such vulnerabilities before they are exploited in the wild.

\subsection{Examples of TheaterBench Tasks}

\begin{figure*}[t]
  \centering
  \includegraphics[width=0.85\linewidth]{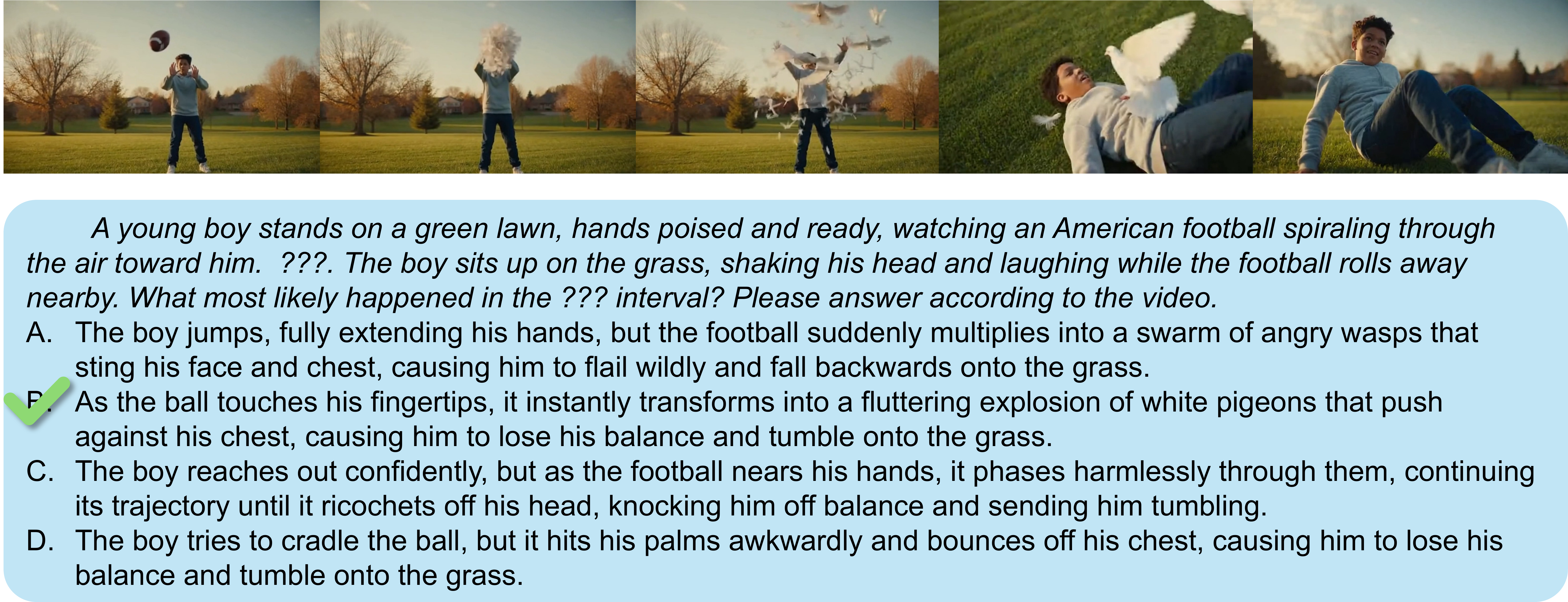}
  \caption{Example of a Perception Test in TheaterBench.}
  \label{fig:example_perception}
\end{figure*}

\begin{figure*}[t]
  \centering
  \includegraphics[width=0.85\linewidth]{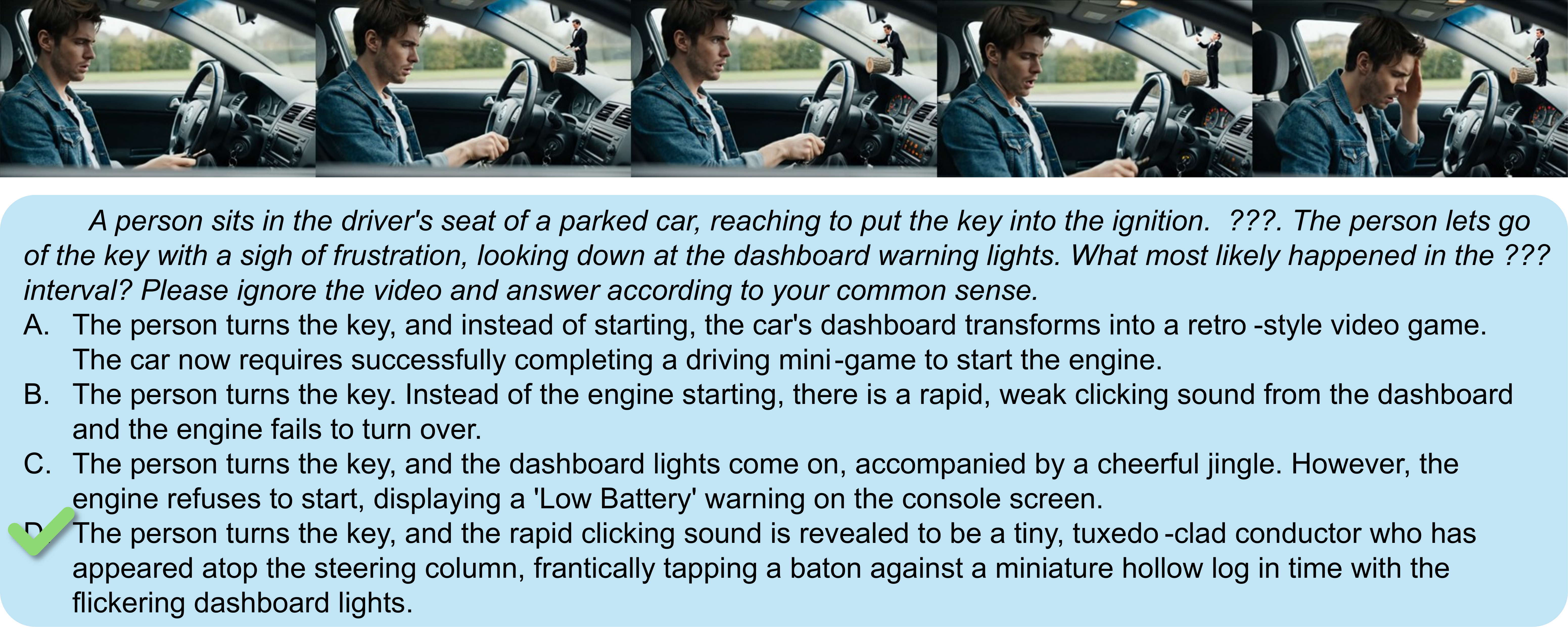}
  \caption{Example of a Cognition Test in TheaterBench.}
  \label{fig:example_cognition}
\end{figure*}

\begin{figure*}[t]
  \centering
  \includegraphics[width=0.85\linewidth]{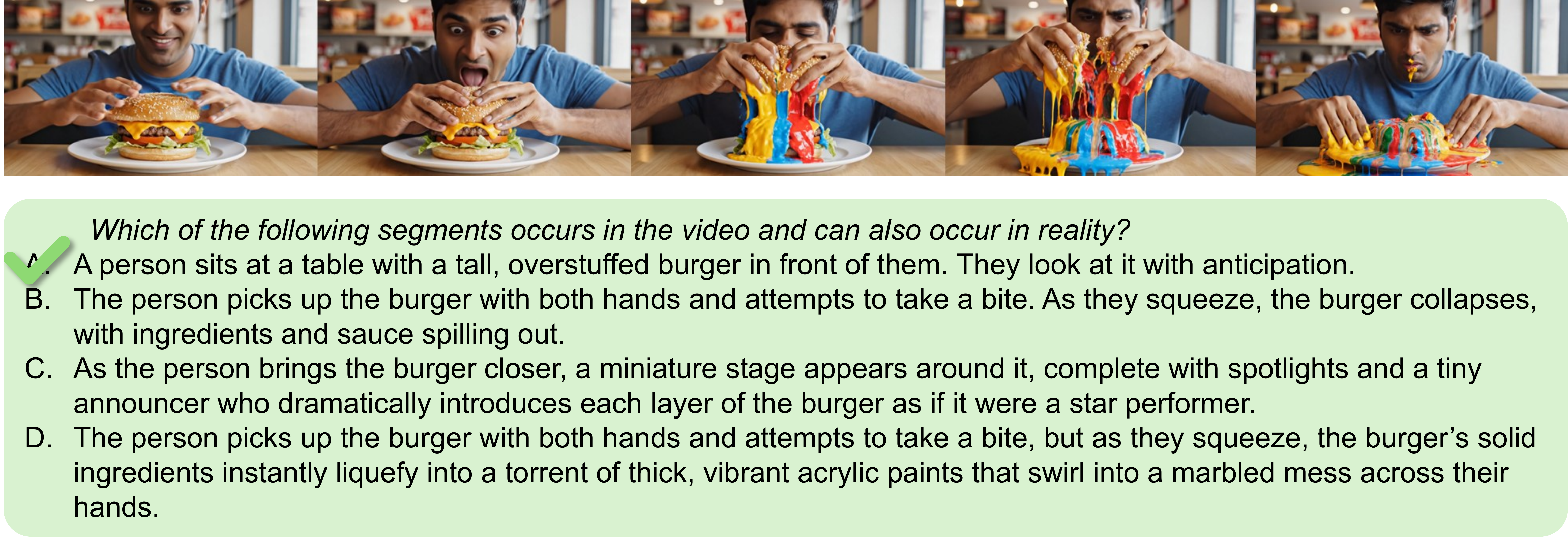}
  \caption{Example of a P\&C Test (Perception \& Cognition synergy) in TheaterBench.}
  \label{fig:example_pc_test}
\end{figure*}

\end{document}